\documentclass[lettersize,journal]{IEEEtran}
\usepackage{amsmath,amsfonts}
\usepackage{algorithmic}
\usepackage{algorithm}
\usepackage{array}
\ifCLASSOPTIONcompsoc
\usepackage{subfig}
\else
\usepackage[caption=false, font=footnotesize]{subfig}
\fi

\usepackage{textcomp}
\usepackage{stfloats}
\usepackage{url}
\usepackage{verbatim}
\usepackage{graphicx}
\usepackage{cite}
\usepackage{booktabs}
\usepackage{hyperref}
\usepackage[hyphenbreaks]{breakurl}

\def \NOTE [#1]{\textcolor{blue}{(\textit{#1})}}
\usepackage{xspace}
\usepackage{color}
\usepackage{multirow}
\usepackage{ifthen}

\long\def\ignorethis#1{}

\definecolor{gray}{rgb}{0.35,0.35,0.35}
\definecolor{MyBlue}{rgb}{0,0.2,0.8}
\definecolor{MyRed}{rgb}{0.8,0.2,0}
\definecolor{MyGreen}{rgb}{0.0,0.5,0.1}
\definecolor{MyGray}{rgb}{0.4,0.4,0.4}

\newlength\paramargin
\newlength\figmargin
\newlength\subfigmargin
\newlength\secmargin
\newlength\subsecmargin
\newlength\tabmargin
\newlength\eqmargin

\usepackage{array}
\newcolumntype{L}[1]{>{\raggedright\let\newline\\\arraybackslash\hspace{0pt}}m{#1}}
\newcolumntype{C}[1]{>{\centering\let\newline\\\arraybackslash\hspace{0pt}}m{#1}}
\newcolumntype{R}[1]{>{\raggedleft\let\newline\\\arraybackslash\hspace{0pt}}m{#1}}

\def\ie{i.e.,~}
\def\eg{e.g.,~}
\def\etc{etc}

\def\vs{vs.~}
\def\etal{et~al.\xspace}

\newcommand{\secref}[1]{Section~\ref{sec:#1}}

\newcommand{\figref}[1]{Fig.~\ref{fig:#1}}
\newcommand{\tabref}[1]{Table~\ref{tab:#1}}
\newcommand{\eqnref}[1]{Eq.\eqref{eq:#1}}

\newcommand{\Paragraph}[1]{\noindent\textbf{#1}}

\usepackage{pifont}
\usepackage{bbding}
\newcommand{\cmark}{\ding{51}}%
\usepackage{float}
\begin{document}

\title{Scalable Face Image Coding via StyleGAN Prior: Towards Compression for Human-Machine Collaborative Vision
}
\def\etal{\textit{et~al}.\xspace}
\def\ie{\textit{i.e.},\xspace}

%\author{Qi Mao, Chongyu Wang, Meng Wang, Shiqi Wang,  Ruijie Chen, Libiao Jin, Siwei Ma}
%\thanks{Qi Mao is with State Key Laboratory of Media Convergence and Communication, Communication University of China, Beijing 100024, China (e-mail: qimao@cuc.edu.cn)}}

% \thanks{J. Doe and J. Doe are with Anonymous University.}% <-this % stops a space
% \thanks{Manuscript received April 19, 2005; revised August 26, 2015.}}
\author{Qi Mao,~\IEEEmembership{Member,~IEEE}, Chongyu Wang, Meng Wang,~\IEEEmembership{Member,~IEEE}, Shiqi Wang,~\IEEEmembership{Senior Member,~IEEE},  Ruijie Chen, Libiao Jin,~\IEEEmembership{Member,~IEEE}, and Siwei Ma,~\IEEEmembership{Fellow,~IEEE}
\thanks{
This work was supported in part by the National Natural Science Foundation of China under Grant 62201526,  Grant 62025101, and Grant 62071449; in part by the National Key Research and Development Project of China under Grant 2021YFF0900502 and 2021YFF0900701; in part by the Fundamental Research Funds for the Central Universities (CUC23GZ007); and in part
by  Public Computing Cloud, CUC, which
are gratefully acknowledged. 
(Corresponding authors: Siwei Ma; Libiao Jin.)

Qi Mao, Chongyu Wang, Ruijie Chen and Libiao Jin are with the School of Information and Communication Engineering and the State Key Laboratory of Media Convergence and Communication, Communication University of China, Beijing 100024, China (E-mail: $\{$qimao, wcy623, chenruijie, libiao$\}$@cuc.edu.cn).

Meng Wang and Shiqi Wang are with the Department of Computer Science, City University
of Hong Kong, Hong Kong, China (E-mail: mwang98-c@my.cityu.edu.hk; shiqwang@cityu.edu.hk).

Siwei Ma is with the National Engineering Research Center of Visual Technology, School of Computer Science,
Peking University, Beijing 100871, China (E-mail:swma@pku.edu.cn).
}}
%

\markboth{Journal of \LaTeX\ Class Files,~Vol.~14, No.~8, August~2021}%
{Qi Mao\MakeLowercase{\textit{et al.}}: A Sample Article Using IEEEtran.cls for IEEE Journals}


\maketitle

\begin{abstract}
The accelerated proliferation of visual content and the rapid development of machine vision technologies bring significant challenges in delivering visual data on a gigantic scale, which shall be effectively represented to satisfy both human and machine requirements.
In this work, we investigate how hierarchical representations derived from the advanced generative prior facilitate constructing an efficient scalable coding paradigm for human-machine collaborative vision.
Our key insight is that by exploiting the StyleGAN prior, we can learn three-layered representations encoding hierarchical semantics, which are elaborately designed into the basic, middle, and enhanced layers, supporting machine intelligence and human visual perception in a progressive fashion.
With the aim of achieving efficient compression, we propose the layer-wise scalable entropy transformer to reduce the redundancy between layers.
Based on the multi-task scalable rate-distortion objective, the proposed scheme is jointly optimized to achieve optimal machine analysis performance, human perception experience, and compression ratio.
We validate the proposed paradigm's feasibility in face image compression.
Extensive qualitative and quantitative experimental results demonstrate the superiority of the proposed paradigm over the latest compression standard Versatile Video Coding (VVC) in terms of both machine analysis as well as human perception at extremely low bitrates ($<0.01$ bpp), offering new insights for human-machine collaborative compression.
\iffalse

\fi
\end{abstract}


\IEEEpeerreviewmaketitle

\section{Introduction}
%Paragraph1-logical flow:
%1.Multimedia increase, data increase, image/video coding is a relatively basic research problem.
%
\IEEEPARstart{R}{ecent} years have witnessed an exponential increase in the amount of image/video data due to the rapid development of various multimedia applications.
Consequently, the highly efficient compression of images and videos has remained a fundamental challenge in multimedia communication and processing for decades.
Early on, images and videos were primarily intended for human viewing and entertainment.
As machine vision technologies advance, growing visual data are required to analyze for intelligent applications, imposing new challenges to machine vision-oriented data compression.
% %%%%%%%%%%%%%%%%%%%%%%%%%%%%%%%%%%%%%%%%%%%%%%%%%%%%%%%
% \begin{figure}[!t]
% \centering
% \includegraphics[width=1\linewidth]{Image/teaser.pdf}
% \\
% %\mpage{0.2}{Input}\hfill\mpage{0.2}{Layer1
% %}\hfill\mpage{0.2}{Layer2}\hfill\mpage{0.2}{Layer3}
% \caption{\textbf{The proposed three-layered scalable face image coding towards human-machine collaborative compression.}
% %
% (a) Input;
% (b) The basic layer which can perform facial landmark detection and segmentation by restoring the sufficient pose, expression, and shape information.
% (c) The middle layer which reconstructs more identity attributes for identity recognition and attribute prediction.
% (d) The enhanced layer which recovers low-level details for human-favored perception.
% %
% The black boxes at the bottom indicate each decoded image's bit rate (bits per pixel/bpp).
%     \label{fig:teaser}
%     \vspace{-5mm}
% \end{figure}
% %%%%%%%%%%%%%%%%%%%%%%%%%%%%%%%%%%%%%%%%%%%%%%%%%%% 
%
The demands of human vision and machine analysis in terms of compression differ fundamentally.
The traditional image compression paradigm for human vision aims to maintain signal fidelity as much as possible under the constraint of the bit rate budget.
In machine vision, retaining and compressing compact features that contain sufficient semantic information for the associated analysis task is commonly practiced.
Both above coding paradigms are well-suited to one vision only but not the other.
In particular, the image compression paradigm cannot guarantee the preservation of semantic information of specific tasks in low-bitrate coding scenarios, which compromises machine analysis efficiency. 
Despite the compact feature being sufficient to support the corresponding vision task, it cannot be reconstructed into visual signals due to the large amount of information lost.
Accordingly, a universal compression scheme that can well serve both human and machine visions is highly desirable~\cite{duan2020video}.

Scalable coding, whose bitstream can be partially decoded for machine analysis and entirely decoded for signal reconstruction, is a natural approach for designing such a human-machine collaborative compression paradigm.
With numerous existing approaches\cite{wang2021towards,yan2021sssic,tu2021semantic,liu2021semantics,hu2020towards,yang2021towards,choi2022scalable} that decompose images into scalable layers, a natural question arises: \textit{What makes one paradigm more efficient than another?}
We assume that scalable layers should support vision tasks in a progressive manner, from simple to complicated, and each layer consumes minimal bitrates to reach the saturated performance.
%

%The recent state-of-the-art StyleGANs, have demonstrated its layer-wise style vectors containing hierarchical semantic information to manipulate specific semantic attributes of images~\cite{shen2020interfacegan,xu2021generative}. 
\IEEEpubidadjcol
In this work, we aim to construct such an efficient scalable coding paradigm for human-machine collaborative vision based on the StyleGAN prior.
Recently, the state-of-the-art StyleGANs~\cite{karras2019style,karras2020analyzing} have demonstrated their layer-wise style vectors containing hierarchical semantic information, which can manipulate specific semantic attributes of synthesized images~\cite{shen2020interfacegan,xu2021generative}.
As such, the layer-wise style vectors can be appropriate representations to control the scalability of reconstructed images serving different purposes.
However, there pose two challenges.
Firstly, each scalable layer should contain \emph{sufficient} semantic representations to accomplish the specific type of vision task.
Furthermore, the cross-layer correlation should be minimized for efficient compression. 
%
%

\begin{figure*}[t]
    \centering
    \includegraphics[width=0.98\linewidth]{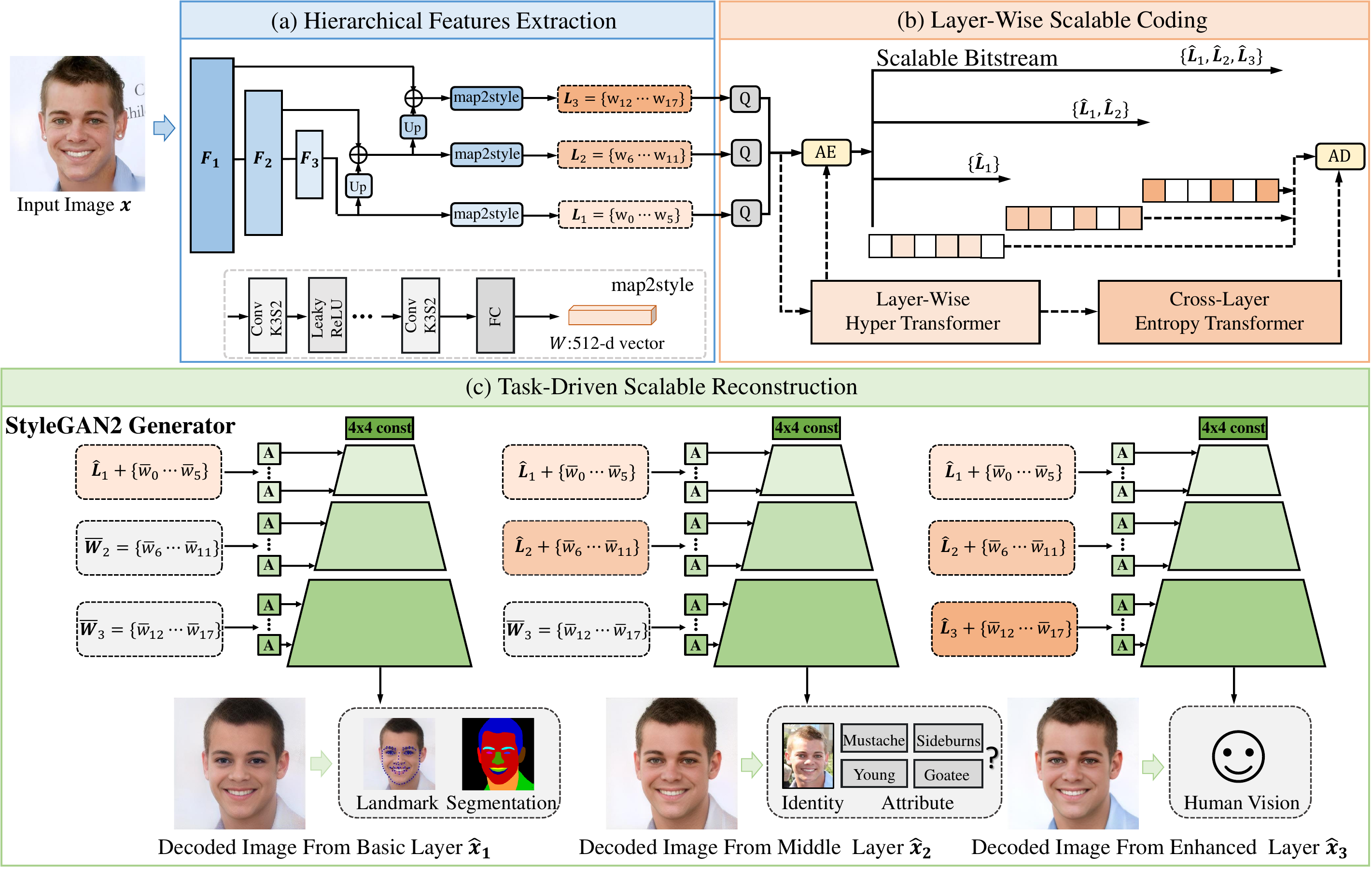}    \caption{\textbf{Overview of the proposed three-layered scalable face image coding framework for human-machine collaborative vision.} (a) Hierarchical three-layered style vectors extraction (\secref{hierarchical-features}); (b) Layer-wise scalable entropy transformer for layer-wise scalable coding (\secref{entropy-model}); (c) Task-driven scalable reconstructions for the basic, middle and enhanced layer via StyleGAN2 generator.
    $\mathbf{F}_1$, $\mathbf{F}_2$, and $\mathbf{F}_3$ represent three levels of feature maps.
$\mathbf{L}_{\ast}$ and $\mathbf{\hat{L}}_{\ast}$ denote the encoded and decoded layer of style vectors $\mathbf{w}_\ast$,
respectively.
$\mathbf{\bar{w}}$ signifies the average style vector of the pre-trained generator.
We use $\mathbf{Q}$, $\mathbf{AE}$, $\mathbf{AD}$ to denote quantization, arithmetic encoding and decoding, respectively.}
    \label{fig:framework}
\end{figure*}

%\IEEEpubidadjcol
To address the challenges mentioned above, we first propose a hierarchical style encoder to divide the layer-wise representations into the basic, middle, and enhanced layers.
Here, we focus on a specific compression
domain, \ie human face images, with significant demand for human-machine collaborative vision in applications such as surveillance and video conference.
As illustrated in~\figref{framework}, the basic layer learns essential contour information, including facial pose, expression, and shape, to perform landmark detection and segmentation tasks.
The middle layer conveys additional information about identity attributes supporting the recognition of identity and attributes analysis, and the enhanced layer includes extra information on low-level signal details aiming at a human-favored reconstruction.
Consequently, the three-layered style vectors are capable of containing sufficient information to progressively serve three types of visual tasks, from simple to complicated. 
To reduce the cross-layer redundancy, we introduce a layer-wise hyper-transformer that employs multi-head attention mechanisms for the hyperprior derivation.
Furthermore, the proposed cross-layer entropy transformer facilitates higher compression ratio by performing conditional probability estimation for the middle and enhanced layers.
Finally, we optimize the entire scheme by the proposed scalable multi-task rate-distortion (R-D) objective, thereby achieving the optimal multi-task performance for machine vision analysis, human perceptual quality, and compression performance.
%

%
%
%%%%%%%%%%%%%%%%%%%%%%%%%%%%%%%%%%%%
%%%%%%%%%%%%%%%%%%%%%%%%%%%%%%%%%%%%%

The main contributions of this work are summarized as follows:
\begin{itemize}
    \item We make one of the first attempts to assign the hierarchical semantic information derived from the StyleGAN prior into the basic, middle, and enhanced layer, which constructs an efficient three-layered scalable face coding framework to progressively support machine and human vision tasks.
    \item We propose a layer-wise hyper-transformer and cross-layer entropy transformer to reduce the correlation between layers for highly efficient compression.
    Furthermore, the scalable multi-task R-D objective is proposed to achieve the optimal multi-task performance on machine vision analysis, perceptual quality, and compression ratio.
    
    \item Extensive vision tasks, including facial landmark detection, parsing, identity recognition, attribute prediction, and human perceptual quality are evaluated on the proposed scheme.
    Both quantitative and qualitative experimental results demonstrate significant performance improvements on each layer over existing codecs, \eg VVC, in terms of corresponding vision tasks, at extremely low bitrates ($< 0.01$ bpp) on high-resolution ($1024 \times 1024$) facial images.
\end{itemize}

The rest of the paper is organized as follows. 
\secref{related work} presents the related work and technologies. 
The proposed framework is illustrated in \secref{proposed-method}. 
\secref{experiments} demonstrates the detailed experimental results and analysis.
Finally, \secref{conclusions} concludes the paper.

\section{Related works}
\label{sec:related work}
\subsection{Image/Video Compression}
Image/video compression technologies aim to compactly encode visual data within a given bitrate constraint, which have been studied extensively for visual communication and processing.
Traditional image/video coding uses a block-based hybrid coding framework that incorporates several handcrafted techniques, \eg prediction, transform, and entropy coding.
There have been numerous image and video coding standards developed over the last several decades, including JPEG~\cite{pennebaker1992jpeg}, JPEG2000~\cite{rabbani2002jpeg2000}, AVC/H.264~\cite{wiegand2003overview}, AVS~\cite{gao2014overview} and HEVC~\cite{sullivan2012overview}.
The latest VVC standard~\cite{bross2021overview} offers more flexible coding techniques to further enhance compression efficiency.
However, the exponential increase in complexity and the rising cost of manually designed tools in existing frameworks pose a challenge to improving compression performance for traditional image and video coding frameworks.

%learning-based method
Thanks to the data-driven and end-to-end (E2E) R-D optimization process, deep neural network-based (DNN-based) image compression methods have demonstrated superior performance to conventional image compression techniques.
% 
% add some new citations
Most mainstream frameworks for deep image compression employ non-linear transform, such as CNN-based architecture~\cite{balle2017end,minnen2018joint,cheng2020image}, to encode an image into a compressed tensor and utilize entropy estimation techniques to estimate bitrates.
Combining more efficient network architectures with more accurate entropy estimation modules~\cite{balle2018variational,minnen2018joint,cheng2020image} makes it possible to achieve a higher compression performance over traditional codecs.
%
%
%GAN-based
Generative adversarial networks (GANs)~\cite{goodfellow2014generative} have also demonstrated its remarkable abilities to reconstruct perceptual-pleasing decoded images under low bitrates in recent compression frameworks~\cite{agustsson2019generative,mentzer2020high,chang2019layered,chang2022conceptual,wang2022disentangled}.
%
%Furthermore, the adversarial training of GAN have shown its impressive capability in recent compression framework to reconstruct the perceptual-pleasing decoded images that much preferred by human vision under low bitrates.
%
%need refine
%
However, all of the frameworks are designed for human vision, which forms the \textit{Compress-then-Analyze} (CTA) paradigm~\cite{redondi2013compress} for machine intelligence applications.
As a result, low bit rate scenarios could result in severely degraded analysis performance due to the highly degradation of semantics.

\subsection{Human-Machine Collaborative Compression}
In light of the growing advances in machine vision algorithms, image/video data analysis has become an essential requirement.
The \textit{Analyze-then-Compress} (ATC) paradigm~\cite{redondi2013compress} is thus proposed by extracting the compact
representation of the visual information~\cite{duan2015overview,duan2018compact}, which achieves lower transmission bandwidth and higher analysis performance.
Nevertheless, such compact
representations cannot be reconstructed into visual signals since the large amount of signal information are lost.

Nowadays, both humans and machines can serve as ultimate visual data receivers.
As such, a new coding paradigm, \ie human-machine collaborative compression~\cite{duan2020video}, has been explored to fulfill the need for both human visual perception and machine visual analysis.
Comparatively, human vision requires more information to reconstruct images at the pixel level. Machine vision, on the other hand, only requires task-specific semantic information for analysis.
Scalable coding, which compresses data into bitstreams that can be partially decoded for different purposes, is a natural design strategy for human-machine collaboration.

Existing research has studied how to decompose images into layers in a scalable manner so that both human and machine vision can be served.
Wang~\etal~\cite{wang2021towards} propose a two-layered scalable framework.
The base layer encodes compact features derived from a pre-trained face recognition model, and the enhancement layer then extracts the residuals of the coarse reconstructed image and the original one for higher-quality reconstruction.
Hu~\etal~\cite{hu2020towards} decompose images into edge maps and color information.
The basic layer employs the compact edge map to reconstruct images for machine vision tasks, while the enhancement layer uses color information to restore more perceptual-consistent images.
Yang~\etal~\cite{yang2021towards} further improve the scalability of the color code by establishing a smooth transition between machine and human vision with the single decoder.
Instead of using independent encoders, Choi \etal~\cite{choi2022scalable} directly split a single latent space into several independent latent features to support machine vision and human vision tasks.
Several recent works~\cite{yan2021sssic,tu2021semantic,liu2021semantics} introduce a ``semantic-to-signal'' concept by representing an image to hierarchical structural features, a certain portion of which are for specific machine analysis tasks, and the entire for pixel-level reconstruction.

However, the semantic features of most existing methods can only handle a single machine vision analysis task, which limits their application scope beyond that. 
In contrast, our proposed scheme's basic and middle layers of decoded images can be applied to a certain range of machine vision tasks.
Additionally, compared with a two-layered edge and color information extraction in~\cite{hu2020towards,yang2021towards}, we leverage hierarchical representations derived from the StyleGAN prior, which performs machine analysis and human perception tasks in an incremental fashion, achieving a more efficient assignment of semantic information.

\subsection{StyleGAN Inversion}
%here
In recent years, GANs~\cite{goodfellow2014generative} have been widely used for generating diverse and realistic images.
The developments of network architectures~\cite{karras2018progressive,karras2019style,karras2020analyzing,brock2018large} and loss functions~\cite{arjovsky2017wasserstein,mao2017least,gulrajani2017improved,mao2019mode} have resulted in significant improvements in the quality and resolution of synthesized images.
Recently, the GAN inversion task~\cite{xia2022gan} is emerging, which derives latent representations of the GAN pre-trained model to reconstruct original real images.

The StyleGAN model~\cite{karras2019style,karras2020analyzing} is one of the most widely used priors for GAN inversion~\cite{abdal2019image2stylegan,abdal2020image2stylegan++}.
Specifically, the generator of StyleGAN can be injected with various latent style vectors at different layers, resulting in diverse styles in the final output.
Therefore, the style vectors derived from the GAN inversion methods~\cite{abdal2019image2stylegan,abdal2020image2stylegan++,richardson2021encoding,tov2021designing} contain semantically rich information, which can manipulate specific semantic attributes of real images~\cite{shen2020interfacegan,xu2021generative}.
In this work, we apply the spirit of StyleGAN inversion to face image compression, which facilitates deriving hierarchical semantic representations to build our efficient scalable coding
paradigm.
%
%In particular, we tailor a hierarchical style encoder to extract three-level semantic representations,
%and leverage the StyleGAN2 generator as a decoder to reconstruct image in a scalable manner.
%

% 
\section{Scalable Face Image Coding Scheme via StyleGAN Prior}
\label{sec:proposed-method}
% Describe the process in general 
%logical flow
%1.Our goal
%2.to achieve our goal, our design
%3.explain the detail of our design 

%
In this work, we aim to develop an efficient scalable face image compression scheme capable of progressively supporting machine analysis and human perception.
One type of machine vision task can be effectively performed with the reconstructed image decoded with a specific set of representations.
By transmitting an additional set of representations, the reconstructed image can support a more complex machine vision task.
Once all representations are obtained, users finally receive a high perceptual quality image.
In order to accomplish this, we utilize the StyleGAN prior, which allows us to control the synthesized images using layer-wise style vectors.
%
%Specific style vectors correspond to particular levels of style during the reconstruction of an image.
%

\figref{framework} presents the overview of the proposed scalable face image coding framework.
In \secref{hierarchical-features}, we first train a hierarchical style encoder to extract $18$ style vectors in accordance with the layer-wise input of the generator in StyleGAN2.
The above style vectors can then be grouped into three layers based on their semantics, aiming to support three levels of vision tasks.
Three-layered representations are sequentially encoded on the encoder side, and on the decoder side, scalable reconstruction effects can be achieved using a StyleGAN2 generator, which constructs the basic, middle and enhanced layer respectively.
The layer-wise scalable entropy transformer model is further proposed in \secref{entropy-model} to fully exploit the redundancy between different layers of style vectors.
With the proposed multi-task scalable R-D objective in \secref{multi-task}, we can optimize all three levels of scalable decoded images to maximize the efficiency of machine vision analysis, human perceptual quality, and compression ratio.

\subsection{Hierarchical Semantic Representations via Style Vectors}
\label{sec:hierarchical-features}
%logical flow
% 
%  
StyleGANs\cite{karras2019style,karras2020analyzing} apply hierarchical layered style vectors to synthesize images. 
In particular, a random noised vector from a prior standard normal distribution $\mathbf{z}\sim \mathcal{Z}$ is first mapped into a 512-dimensional vector $\mathbf{w}\sim \mathcal{W}$ through a learnable fully-connected mapping network.
Various affine transforms are then applied to the $\mathbf{w}$ vector to control output styles at different resolutions.
%The $\mathbf{w}$ vector is then fed into different affine transforms of the generator to control the output styles at different resolutions.
%
Generally, the style vectors correspond to distinct styles that can be roughly grouped into coarse, middle, and high-level~\cite{karras2019style}.
Consider the face image as an example: 
Style vectors at the coarse level can affect an image's pose, expression, and face shape.
The middle level of style vectors contains more facial details, and the fine level represents details such as color and microstructure.
In light of these observations, we embed the input images into these hierarchical style vectors and group them into three layers.

We first consider embedding the input image $\mathbf{x} \in \mathcal{R}^{1024\times1024\times3}$ into the $\mathcal{W}$ space. 
However, using the same 512-dimensional vectors~\cite{xia2022gan} has been demonstrated to yield inaccurate reconstructions.
As a common practice, images
are embedded into an extended latent space, $\mathbf{w} \in \mathcal{W}^{+}$, that consists of different 512-dimensional $\mathbf{w}$ vectors, \eg $18$ different styles for generating $1024\times1024$ images.
Therefore, we design a hierarchical style encoder $E$ to extract $18$ different style vectors as,
\begin{equation}
    E(\mathbf{x}) = \{\mathbf{w}_0,\cdots,\mathbf{w}_5,\cdots,\mathbf{w}_{11},\cdots,\mathbf{w}_{17} \}.
\end{equation}
% and group them accordingly as,
% \begin{equation}
%  \small
%  \begin{aligned}
%     \mathbf{L} &= E(\mathbf{x}), \\
%   \mathbf{L} &=\{\mathbf{L}_1, \mathbf{L}_2, \mathbf{L}_3\}.\\
% \end{aligned}
% \end{equation}
%
%

In particular, as presented in \figref{framework}(a), we extract three levels of feature maps $\mathbf{F}=\{\mathbf{F}_1, \mathbf{F}_2, \mathbf{F}_3\}$ in different resolutions and then adopt the map2style block in \cite{richardson2021encoding} to equally map each feature into $6$ style vectors, 
\ie the small feature map extracts the first layer of style vectors $\mathbf{L}_1=\{\mathbf{w}_0,\cdots,\mathbf{w}_5\}$; By combining the upsampling small feature map, the medium feature map extracts the second layer of style vectors $\mathbf{L}_2=\{\mathbf{w}_6,\cdots,\mathbf{w}_{11}\}$; While the largest feature map, which fuses the up-sampling medium feature map, extracts the third layer of style vectors $\mathbf{L}_3=\{\mathbf{w}_{12},\cdots,\mathbf{w}_{17}\}$.
The $18$ style vectors thus are grouped accordingly as,
\begin{equation}
 \mathbf{L} =\{\mathbf{L}_1, \mathbf{L}_2, \mathbf{L}_3\}.
\end{equation}
%

%Therefore, the StyleGAN2 reconstructs the decoded image on the decoder side based on the transmitted style vectors.
%
Similar to \cite{richardson2021encoding}, we introduce the average style vector $\mathbf{\bar{w}}$ of the pre-trained StyleGAN2 generator $G$ and reconstruct the image $\hat{\mathbf{x}}$ as,
\begin{equation}
    \hat{\mathbf{x}} = G(E(\mathbf{x}) +\mathbf{\bar{w}}).
\end{equation}
%where $E(\cdot)$ and $G(\cdot)$ denote the proposed hierarchical style encoder and StyleGAN2 generator, respectively. 
%

The encoder is thus forced to learn different \textbf{residue style vectors} against the average style vector, which are more sparse and easier to compress.
Then, the disentangled representations allow us to gradually transmit layer-wise style vectors, as shown in \figref{framework}(b).
Consequently, the StyleGAN2 generator reconstructs the decoded image based on the transmitted style vectors.
\figref{framework}(c) illustrates the three-layer decoded images $\{\hat{\mathbf{x}}_{1}, \hat{\mathbf{x}}_{2},\hat{\mathbf{x}}_{3}\}$ using scalable biterates to support three levels of vision tasks.

When only the first layer of style vectors is transmitted, we expect the partially decoded image from \textbf{the basic layer} $\hat{\mathbf{x}}_{1}$ to maintain the basic contour such as facial pose, expression, and shape of the original image, which can be used for tasks such as landmark detection and face parsing.
\begin{equation}
    \hat{\mathbf{x}}_{1} = G(\{\mathbf{{L}}_1,0,0\}+\mathbf{\bar{w}}).
\end{equation}

Upon receiving the second level of style vectors, the reconstructed image from \textbf{the middle layer}  $\hat{\mathbf{x}}_{2}$ is able to recover the original image's semantic attributes, facilitating more complex tasks such as facial identity recognition and attribute prediction. 
\begin{equation}
    \hat{\mathbf{x}}_{2} = G(\{\mathbf{{L}}_1,\mathbf{{L}}_2,0\}+\mathbf{\bar{w}}).
\end{equation}

In the end, the final decoded image from \textbf{the enhanced layer}  $\hat{\mathbf{x}}_{3}$ preferred by humans is obtained once all the style vectors have been transmitted.
\begin{equation}
    \hat{\mathbf{x}}_{3} = G(\{\mathbf{{L}}_1,\mathbf{{L}}_2,\mathbf{{L}}_3\}+\mathbf{\bar{w}}).
\end{equation}

\subsection{Layer-Wise Scalable Entropy Transformer Model}
\label{sec:entropy-model}
%logical flow 
%1.to estimate the rate.
%2.how to estimate, what redundancy to exploit

%By extracting the compact style vectors in Section\ref{hierarchical-features}, all the style details can be captured.
%
In order to add rate constraints for E2E R-D optimization, we use the entropy model to estimate the probability distribution of style vectors.
In contrast to E2E image compression frameworks\cite{balle2017end,balle2018variational}, this work eliminates layer-wise redundancy of style vectors instead of spatial redundancy.

%
%1.how to use hyperpior
%
Following the spirit of \cite{balle2018variational}, we introduce the hyperprior model to capture the correlation of layered style vectors.
Each style vector $\mathbf{w}_i$ is first quantized into $\hat{\mathbf{w}}_i = Q(\mathbf{w}_i)$.
Due to the non-differentiable of quantization operation, the uniform noise\cite{balle2017end} is added during training capable of back-propagation for E2E training.
Then each quantized style vector $\hat{\mathbf{w}}_i$ is estimated by a conditional Guassian model, where the mean $\mathbf{\mu}_i$ and the variance $\mathbf{\delta}_i$ is derived from the quantized hyperprior $\hat{\mathbf{h}}_i=Q(\mathbf{h}_i)$ as:
\begin{equation}
   p_{\hat{\mathbf{w}}_i|\hat{\mathbf{h}}_i}(\hat{\mathbf{w}}_i|\hat{\mathbf{h}}_i) = \mathcal{N}(\mathbf{\mu}_i, \mathbf{\delta}_i^2).
\end{equation}

The probability of the quantized hyperprior $\hat{\mathbf{h}}_i$ is modeled by fully factorized entropy model~\cite{balle2017end}, which are also need to be transmitted as the side information.
Therefore, the rate constraints of three-layered style vectors $\hat{\mathbf{L}}$ are derived as,

\begin{equation}
\label{eq:all}
   \mathcal{R}(\mathbf{\hat{L}}) = \sum_{i=0}^{17}\{\mathbb{E}[-\rm{log}_2{p_{\hat{\mathbf{w}}_i}}(\hat{\mathbf{w}}_i)]+\mathbb{E}[-\rm{log}_2{p_{\hat{\mathbf{h}}_i}}(\hat{\mathbf{h}}_i)]\}.
\end{equation}

%1.the hyperpior model using transformer.
%2.note the masked one
\begin{figure}[t]
    \centering
    \includegraphics[width=1\linewidth]{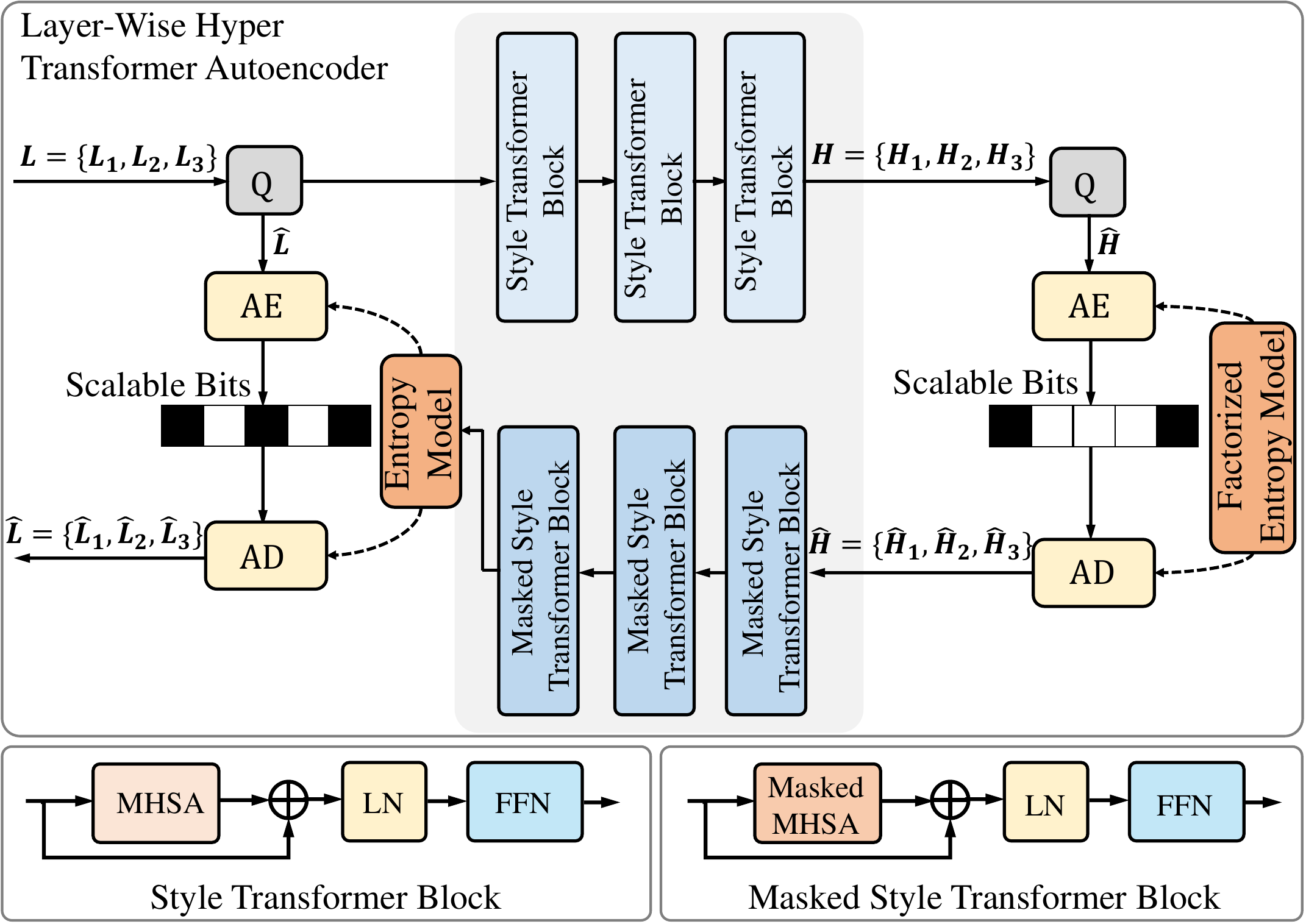}
    \caption{\textbf{The proposed layer-wise hyper-transformer.}
    ``MHSA'' and ``Masked MHSA'' refer to the multi-head self attention and the masked multi-head self-attention, respectively.
    ``LN'' indicates the layer normalization, and ``FFN'' represents the fully-connected feed-forward network.
    %
    %Unsent style vectors are masked in gray color blocks, which are not considered when calculating attention scores.
    }
    \label{fig:layer-hp}
\end{figure}

\Paragraph{Layer-Wise Hyper-Transformer.}
To fully exploit the correlation between style vectors, we propose to reduce the redundancy from two aspects: a) the redundancy between style vectors; and b) the cross-channel redundancy of one style vector $\hat{\mathbf{w}}_i$.
Hence, we propose a layer-wise hyper-transformer autoencoder model that can achieve both goals, consisting of a hyper-encoder and a hyper-decoder, as presented in \figref{layer-hp}.
Following \cite{vaswani2017attention}, the style transformer block is proposed as a basic structure, which includes the multi-head self-attention module, the residual connection, layer normalization, and the fully-connected feed-forward network.
Different quantized style vectors $\hat{\mathbf{w}}_i$ can be regarded as query tokens, denote as $\{q_0, \cdots, q_i, \cdots, q_{17}\}$.
Using the multi-head self-attention module, we can determine the correlation between any pair of style vectors and obtain an attention score.
We denote all of the query tokens as $X_s \in \mathbb{R}^{18\times512}$.
Then, in each head of the self-attention module, the query $Q$, key $K$ and value $V$ are all projected from $X_s$ via learnable projection heads as follows.
\begin{equation}
    Q=X_s W^{self}_{Q}, \quad K=X_s W^{self}_{K}, \quad V=X_s W^{self}_{V},
\end{equation}
where $W^{self}_{Q}$,$W^{self}_{K}$, and $W^{self}_{V} \in \mathbb{R}^{512\times512}$, which do not change the dimensions of style vectors.
The scaled dot-product attention in \cite{vaswani2017attention} then can be calculated as,
\begin{equation}
\label{eq:attn}
    Attn(Q,K,V)=Softmax(\frac{QK^{T}}{\sqrt{d}})V,
\end{equation}
where $d= 512 / H$, and $H$ is the number of attention heads.
We empirically adopt $H=4$ in our experiments.
Therefore, final results combines all attention of different heads can be written as,
\begin{equation}
\label{eq:mha}
MHA(Q,K,V)=[Attn(Q_i,K_i, V_i)]_{i=1:H} W^{o},
\end{equation}
where $Q_i$, $K_i$, and $V_i$ the $i$-th attention head' query, key and value, $W^{o} \in \mathbb{R}^{512\times512}$ denotes an learnable matrix to combine all attention results.
Additionally, the dimension of style vectors is reduced with a fully-connected layer feed-forward network to eliminate cross-channel redundancy.
%Furthermore, in order to eliminate the cross-channel redundancy, we reduce the dimensions of style vectors using fully-connect layer feed-forward net.
%
The derived hyperprior dimensions are reduced from $d=512$ to $d=16$ after three repeated style transformer blocks.

\begin{figure}[!b]
    \centering
    \includegraphics[width=1\linewidth]{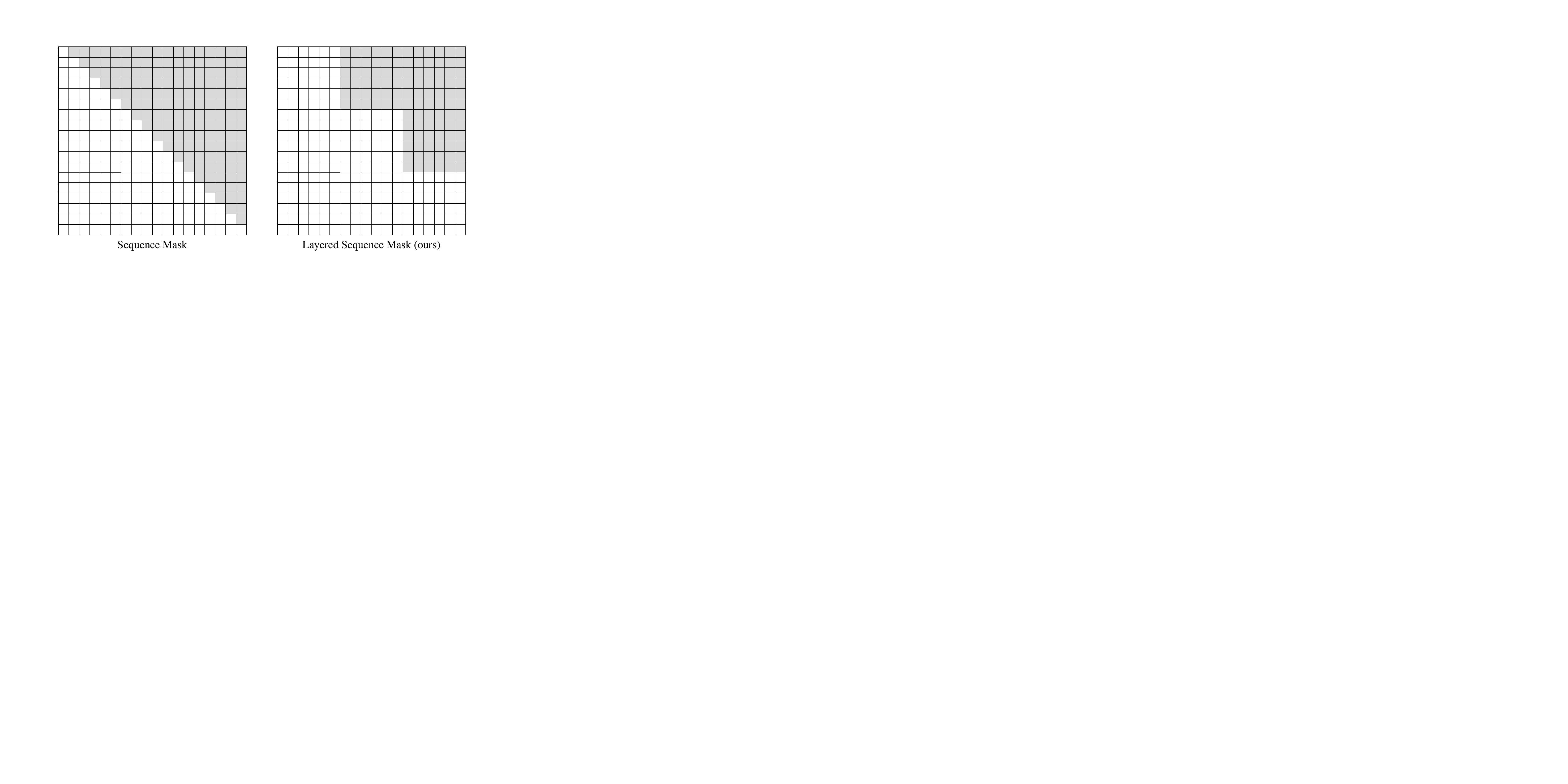}
    \caption{\textbf{Illustration of the proposed layered sequence mask (right) \vs the sequence mask in~\cite{vaswani2017attention} (left).}
    Unsent style vectors are masked in gray color blocks, which are not considered when calculating attention scores.}
    \label{fig:mask}
\end{figure}

Regarding the transmission order of the scalable coding framework, the hyper-transformer decoder can only compute the mean and variance of the current style vector based on the previous and current layers of hyperprior.
To implement the above process during training, we propose a masked multi-head self-attention module and a masked style transformer in the hyper-decoder.
\figref{mask} illustrates how the proposed layered sequence mask differs from the sequence mask of the original transformer~\cite{vaswani2017attention}, which regards three layers of style vector transmission order.
Consequently, it will not consider the hyperprior of unsent layers when calculating attention scores.
Likewise, the masked style transformer block is applied three times and predicts the $\mu$ and $\delta$ for each quantized style vector.

\Paragraph{Cross-Layer Entropy Transformer.}
We can obtain the decoded previous layer of style vectors with the scalable setting when estimating the second and third layers of style vectors.
Thus, the second layer and third layer style vectors' entropy estimation cannot only rely on the information of hyperpriors but also on the decoded style vectors of previous layers.
The \eqnref{all} then can be rewritten as,
\begin{equation}
 \small
 \begin{aligned}
\mathcal{R}(\hat{\mathbf{L}}_1) &=\mathbb{E}[-\rm{log}_2{p_{\hat{\mathbf{L}}_1|\hat{\mathbf{H}}_1}(\hat{\mathbf{L}}_1|\hat{\mathbf{H}}_1})]\\
+& \mathbb{E}[-\rm{log}_2{p_{\hat{\mathbf{H}}_1}}(\hat{\mathbf{H}}_1)], \\
\mathcal{R}(\hat{\mathbf{L}}_2)&=\mathbb{E}[-\rm{log}_2{p_{\hat{\mathbf{L}}_1|\hat{\mathbf{H}}_2, \hat{\mathbf{L}}_1}(\hat{\mathbf{L}}_2|\hat{\mathbf{H}}_1},\hat{\mathbf{L}}_1)] \\
+&\mathbb{E}[-\rm{log}_2{p_{\hat{\mathbf{H}}_2}}(\hat{\mathbf{H}}_2)], \\
\mathcal{R}(\hat{\mathbf{L}}_3)&=\mathbb{E}[-\rm{log}_2{p_{\hat{\mathbf{L}}_1|\hat{\mathbf{H}}_3, \hat{\mathbf{L}}_1, \hat{\mathbf{L}}_2}(\hat{\mathbf{L}}_3|\hat{\mathbf{H}}_3},\hat{\mathbf{L}}_1,\hat{\mathbf{L}}_2)]\\
   +&\mathbb{E}[-\rm{log}_2{p_{\hat{\mathbf{H}}_3}}(\hat{\mathbf{H}}_3)], \\
\end{aligned}
\end{equation}
where $\hat{\mathbf{L}}_i$ and $\hat{\mathbf{H}}_i$ denotes the $i$-th layer of quantized style vectors and hyperpriors, respectively.
By utilizing the decoded previous layers, we propose a cross-layer entropy transformer that improves R-D optimization by estimating the probability of the current layer more accurately.

\begin{figure}[t]
    \centering
    \includegraphics[width=1\linewidth]{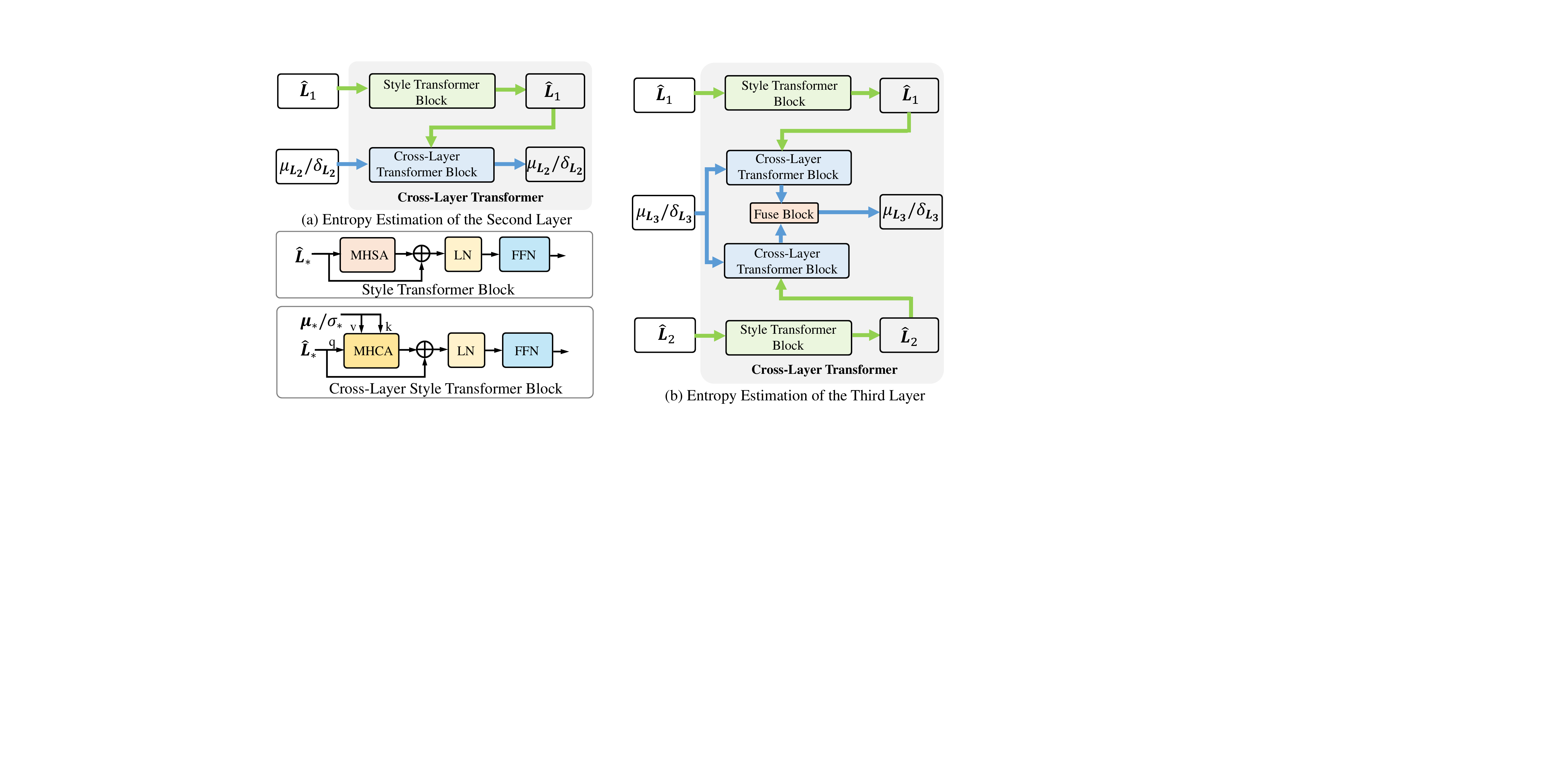}
    \caption{\textbf{The proposed cross-layer entropy transformer.}
    ``MHCA'' indicates the multi-head cross attention.
    ``LN'' and ``FFN'' mean the layer normalization, the fully-connected feed-forward network, respectively. 
    }
    \label{fig:cross-layer}
\end{figure}

As shown in \figref{cross-layer}(a), we take the mean $\mu_{\mathbf{L}_2}$ of second layer as an example, the decoded first layer referred as $X_{\hat{\mathbf{L}}_1} \in \mathbb{R}^{6\times512}$ are first feed into a style transformer block, deriving the updated tokens as $\tilde{X}_{\mathbf{\hat{L}}_1} \in \mathbb{R}^{6\times512}$.
A cross-layer style transformer block consists of a multi-head cross-attention module, a layer normalization, and a feed-forward network, which is then introduced in order to combine the decoded first layer with the $\mu_{\mathbf{L}_2}$ predicted from layer-wise hyper-transformers, denoted as $X_{\mu_{\mathbf{L}_2}} \in \mathbb{R}^{6\times512}$.
%We then introduce a multi-head cross-attention module to combine it with the $\mu_{\mathbf{L}_2}$ predicted from layer-wise hyper-transformer.
%
%
The query $Q$, key $K$ and value $V$ of multi-head cross attention are derived as follows.
\begin{equation}
    Q=X_{\mu_{\mathbf{L}_2}} W^{crs}_{Q},  \quad K=\tilde{X}_{\mathbf{\hat{L}}_1} W^{crs}_{K}, \quad V=\tilde{X}_{\mathbf{\hat{L}}_1} W^{crs}_{V},
\end{equation}
where $W^{crs}_{Q}$, $W^{crs}_{K}$, and $W^{crs}_{V} \in \mathbb{R}^{512\times512}$.
Using the same calculation in \eqnref{attn} and \eqnref{mha}, we compute the cross-attention score for multiple heads.
%We then calculate the multi-head cross attention scores using the same way as \eqnref{attn} and \eqnref{mha}.
%
\figref{cross-layer}(b) demonstrates that the decoded first and second layers' style vectors help derive the last layer predictions.
In particular, we apply three convolutional blocks to construct the fuse block.

%the cross-layer estimation

\subsection{Multi-Task Scalable Rate-Distortion Optimization}
\label{sec:multi-task}
Following human-machine collaborative paradigm~\cite{duan2020video}, our goal is to maximize the both human and machine vision tasks' performance with the constraint of bitrates.
We first construct a R-D objective of each decoded layered image as,
%\begin{equation}
 %\small
\begin{align}
 %\label{eq:RD}
 \label{eq:RD-1}
\mathcal{J}_{\mathbf{\hat{L}}_{1}} & =\lambda \mathcal{R}(\mathbf{\hat{L}}_{1})+ \mathcal{D}^{M_1}(\hat{\mathbf{x}}_{1}),\\ 
\label{eq:RD-2}
\mathcal{J}_{\mathbf{\hat{L}}_{2}} & =\lambda(\mathcal{R}(\mathbf{\hat{L}}_{1})+\mathcal{R}(\mathbf{\hat{L}}_{2}))+ \mathcal{D}^{M_2}(\hat{\mathbf{x}}_{2}),\\
\label{eq:RD-3}
    \mathcal{J}_{\mathbf{\hat{L}}_{3}}&=\lambda (\mathcal{R}(\mathbf{\hat{L}}_{1})+\mathcal{R}(\mathbf{\hat{L}}_{2}) + \mathcal{R}(\mathbf{\hat{L}}_{3})) + \mathcal{D}^{H}(\hat{\mathbf{x}}_{3}),
\end{align}
%\end{equation}
where $\lambda$ indicates the Lagrangian multiplier to control the R-D trade-off.
%$Lagrange$ 
We then develop distortion metrics using three-level vision tasks to optimize machine and human vision performance.
It should be noted that, although several predefined vision tasks are selected for optimization, the reconstructed image can also be applied to other tasks. 

\Paragraph{Landmark Detection Distortion.} To assess the facial landmark detection task distortion, we directly feed the decoded image and the original image to the pre-trained landmark detection network\cite{wang2019adaptive} to measure the distortion of the predicted landmark heatmaps as,
\begin{equation}
    \mathcal{D}^{M}_{lm}(\hat{\mathbf{x}})=\|H(\mathbf{x}) - H(\hat{\mathbf{x}}) \|_2^{2},
\end{equation}
where $H(\cdot)$ represents the landmark heatmap extracted by the pre-trained landmark detection network.
%\lambda_{lm}

\Paragraph{Facial Segmentation Distortion.} We introduce a pre-trained facial parsing network $P$ \cite{CelebAMask-HQ} to measure the distortion between the decoded image and the original image.
Instead of directly measuring the parsing map distortion, we minimize the cosine-distance between multiple layers of features extracted from the pre-trained facial parsing network as,

\begin{equation}
    \mathcal{D}^{M}_{sg}(\hat{\mathbf{x}})=\sum_{i=1}^{5}(1-cos(P_i(\hat{\mathbf{x}}),P_i(\mathbf{x}))),
\end{equation}
where $P_i(\cdot)$ represents the $i$-th feature extracted by the pre-trained facial parsing network, $cos$ means the cosine similarity.
%\lambda_{sg}

\Paragraph{Facial Identity Distortion.} The pre-trained face recognition network (ArcFace)\cite{deng2019arcface} is applied to measure the distortion of identity by minimizing the cosine-distance of multiple features.
\begin{equation}
    \mathcal{D}^{M}_{id}(\hat{\mathbf{x}})=\sum_{i=1}^{5}(1-cos(R_i(\hat{\mathbf{x}}),R_i(\mathbf{x}))),
\end{equation}
where $R_i(\cdot)$ represents the $i$-th feature extracted by the pre-trained face recognition network.
%\lambda_{ID}

\Paragraph{Human Perceptual Distortion.} 
We adopt $L_2$ loss to provide pixel-level supervision and LPIPS loss~\cite{zhang2018unreasonable} to measure perceptual quality as,
%\begin{equation}
%\label{eq:HP}
%\begin{split}
\begin{align}
    \mathcal{D}^{H}_{mse}(\hat{\mathbf{x}})&= \|\mathbf{x} - \hat{\mathbf{x}}\|_2^{2}, \\
    \mathcal{D}^{H}_{\rm{LPIPS}}(\hat{\mathbf{x}})&= \|F(\mathbf{x})-F(\hat{\mathbf{x}}) \|_2^{2}. 
\end{align}
%\end{split}    
%\end{equation}
where $F(\cdot)$ denotes the AlexNet~\cite{krizhevsky2012imagenet} feature extractor.
%\lambda_{\rm{mse}}
%\mathcal{L}_{\rm{LPIPS}}

\Paragraph{Adversarial Training.}
Additionally, we also employ the StyleGAN2 discriminator $D$ to perform the adversarial training $\mathcal{L}_{adv}$, which can further enrich the texture and enhance the realism of decoded images.
Following~\cite{karras2019style,karras2020analyzing}, we adopt the non-saturating adversarial loss \cite{goodfellow2014generative} with $R_1$ regularization \cite{mescheder2018training}.

\begin{equation}
 \begin{aligned}
   \mathop{min}\limits_{E,G}{D}^{H}_{adv}(\hat{\mathbf{x}}) & = \mathcal{L}_{adv}^{E,G}=\log(\exp(-D(\hat{\mathbf{x}}))+1), \\
   \mathop{min}\limits_{D} \mathcal{L}_{adv}^{D}&=\log(\exp(D(\hat{\mathbf{x}})))+1) \\
   +&\log(\exp(-D(\mathbf{x}))+1)+ 0.5\cdot \lambda_{r_1}\|\Delta_{\mathbf{x}}D(\mathbf{x})\|_2^2.
\end{aligned}
\end{equation}
%\lambda_{\rm{adv}}

%
The decoded image from the basic layer maintains the facial pose, expression, and shape information, which can support the facial landmark detection and segmentation tasks, \eqnref{RD-1} thus can be rewritten as,
\begin{equation}
\label{eq:rd-l1}
    \mathcal{J}_{\mathbf{\hat{L}}_{1}}=\lambda \mathcal{R}(\mathbf{\hat{L}}_{1})+ \lambda_{lm} \mathcal{D}^{M}_{lm}(\hat{\mathbf{x}}_{1}) + \lambda_{sg} \mathcal{D}^{M}_{sg}(\hat{\mathbf{x}}_{1}),
\end{equation}
where $\lambda_{\ast}s$ indicate the weighting parameters of losses.

Additionally to the vision tasks that can be carried out with the basic layer, by recovering more facial semantics, the decoded image from the middle layer can be used for facial identity recognition.
Therefore,  \eqnref{RD-2} can be rewritten as,
\begin{equation}
\label{eq:rd-l2}
\begin{split}
    \mathcal{J}_{\mathbf{\hat{L}}_{2}}&=\lambda (\mathcal{R}(\mathbf{\hat{L}}_{1})+\mathcal{R}(\mathbf{\hat{L}}_{2})) +\\ &w_{1} \cdot(\lambda_{lm}\mathcal{D}^{M}_{lm}(\hat{\mathbf{x}}_{2}) +  \lambda_{sg}\mathcal{D}^{M}_{sg}(\hat{\mathbf{x}}_{2})) +\lambda_{ID}\mathcal{D}^{M}_{ID}(\hat{\mathbf{x}}_{2}),
\end{split}    
\end{equation}
where $\lambda_{\ast}s$ are loss weights to balance different loss terms.
With additional bitrates, the performance of vision tasks supported by the basic layer could also be enhanced.
We enhance the scalability of the middle layer's vision performance by strengthening the weighting constraint through the utilization of $w_1$. For our experimental investigations, we set $w_1=1.5$ as the value for this weight.

Finally, the decoded image from the enhanced layer aims to reconstruct for human vision with high perceptual quality.
The \eqnref{RD-3} then can be formulated as,
\begin{equation}
\label{eq:rd-l3}
\begin{split}
    \mathcal{J}_{\mathbf{\hat{L}}_{3}}&=\lambda (\mathcal{R}(\mathbf{\hat{L}}_{1})+\mathcal{R}(\mathbf{\hat{L}}_{2}) + \mathcal{R}(\mathbf{\hat{L}}_{3}))\\ +  &w_{2}\cdot(\lambda_{lm}\mathcal{D}^{M}_{lm}(\hat{\mathbf{x}}_{3}) +\lambda_{sg} \mathcal{D}^{M}_{sg}(\hat{\mathbf{x}}_{3})) \\
    &+ w_{3}\cdot\lambda_{ID}\mathcal{D}^{M}_{ID}(\hat{\mathbf{x}}_{3}) 
    + \lambda_{mse}\mathcal{D}^{H}_{mse}(\hat{\mathbf{x}}_{3}) \\
    & + \lambda_{\rm{LPIPS}}\mathcal{D}^{H}_{\rm{LPIPS}}(\hat{\mathbf{x}}_{3}) +
    \lambda_{\rm{adv}}\mathcal{D}^{H}_{adv}(\hat{\mathbf{x}}_{3}).
\end{split}    
\end{equation}
where the terms $\lambda_{\ast}s$ control the importance of each loss function.
To formulate scalability, we incorporate weights $w_{2}$ and $w_{3}$ to reflect the enhanced performance of vision tasks within the basic and middle layers as the bitrates increase.
With the increase in bitrates, the enhanced layer's performance in landmark detection and segmentation tasks exhibits a more pronounced improvement compared to the middle layer.
Consequently, in our experimental investigation, we assign $w_2=2$ and $w_3=1.5$ as the respective values for these weights.

In the end, the proposed multi-task scalable R-D objective of our compression framework can be written as,
\begin{equation}
    \mathcal{J}_{Scalable}= \mathcal{J}_{\mathbf{\hat{L}}_{1}}+ \mathcal{J}_{\mathbf{\hat{L}}_{2}}+ \mathcal{J}_{\mathbf{\hat{L}}_{3}}.
\end{equation}

\section{Experiments}
\label{sec:experiments}
\subsection{Implementation Details}
\Paragraph{Dataset.}
We train the proposed method on the FFHQ dataset~\cite{karras2019style} consisting of $70,000$ facial images in $1024\times1024$ resolution, and evaluate it on the CelebA-HQ dataset~\cite{karras2018progressive}, which has $30,000$ $1024\times1024$ resolution of images.

\Paragraph{Training Process.}
The proposed model is implemented on the PyTorch~\cite{paszke2017automatic} framework and trained the model on one NVIDIA Tesla-A$100$ GPUs with $40$ GB memory.
We empirically adopt the following hyper-parameters in all the experiments:
$\lambda_{lm}=1$, $\lambda_{sg}=1$, $\lambda_{ID}=0.5$, $\lambda_{{mse}}=1$, $\lambda_{\rm{LPIPS}}=0.8$, $\lambda_{adv}=0.01$, and $\lambda_{r_1}=10$.
To adjust the bitrate range, $\lambda$ belongs to the set $\{5,10,15,20\}$.
We use the batch size of $4$ as well as the Adam \cite{kingma2014adam} optimizer with a learning rate of $10^{-4}$ and exponential decay rates $(\beta_1, \beta_2)=(0, 0.99)$. 
First, the weights are initialized using the pre-trained StyleGAN2 generator and discriminator, and the generator is then optimized simultaneously with the hierarchical style encoder, the layer-wise hyper-transformer, and the cross-layer entropy transformer.
We train the framework without the adversarial training for $50$K iterations, and then we add the adversarial training for another $10$K iterations.

\subsection{Machine Vision: Compression Evaluation on the Basic Layer}

%\subsubsection{Evaluation Tasks}
% 
By reconstructing the image using the first layer of style vectors, the basic layer's decoded image $\hat{\mathbf{x}}_1$ preserves the basic contour information of the original image.
Therefore, we evaluate it on two machine vision tasks, facial landmark detection and face parsing, in terms of task accuracy \vs bitrate.

\begin{figure*}[t]
    \centering
    % \vspace{-3mm}
    \includegraphics[width=0.98\linewidth]{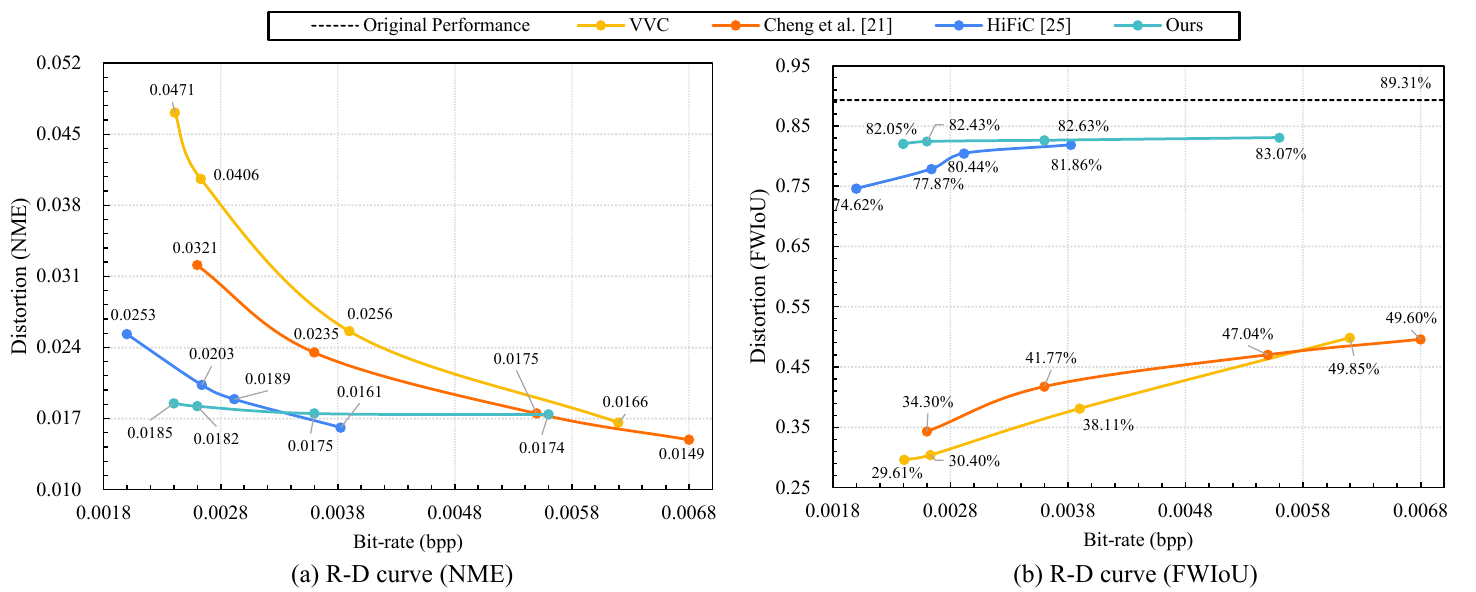}
   \caption{\textbf{The R-D performance of VVC, Cheng \etal~\cite{cheng2020image}, HiFiC~\cite{mentzer2020high}, and the basic layer on the CelebA-HQ dataset for facial landmark detection and parsing. }
   (a) Lower NME values indicate better landmark detection accuracy; 
   (b) Higher FWIoU values mean better parsing accuracy.
   %
   %The proposed method can achieve much better accuracy at similar bit-rate. 
%
}
    \vspace{-4mm}
    \label{fig:RD-layer1}
\end{figure*}
%%%%%%%%%%%%%%%%%%%%%%%%%%%%%%%%%%%%%%%%%%%%%%%%%%%%%%
\begin{figure}[!t]
\centering
\centering
\includegraphics[width=1.0\linewidth]{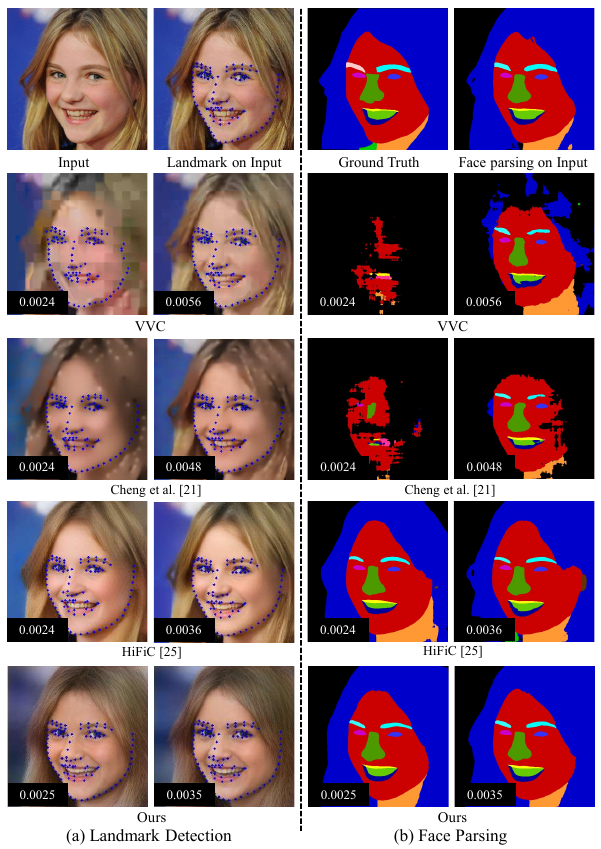}
\caption{\textbf{Visual comparisons with VVC, Cheng \etal~\cite{cheng2020image} , and HiFiC~\cite{mentzer2020high}} on facial landmark detection and parsing.
We show the detected landmarks as blue circles.
Each decoded image's bit rate (bits per pixel/bpp) is shown in the black boxes at the bottom.
  }
    \label{fig:layer1}
\end{figure}
%%%%%%%%%%%%%%%%%%%%%%%%%%%%%%%%%%%%%%%%%%%%%%%%%%%

% 

\Paragraph{Benchmarks.}
We compare the proposed method with both traditional standard and DNN-based compression frameworks.
First, we compare with the latest video coding standard VVC and employ all intra configuration of the reference software VTM-11.0~\footnote{\url{https://vcgit.hhi.fraunhofer.de/jvet/VVCSoftware_VTM/-/tree/VTM-11.0}} using QP from $\{63, 62, 58, 54\}$ to cover the proposed method's bitrate range.
For DNN-based method, we compare with Cheng \etal \cite{cheng2020image} and retrain the model implemented by CompressAI~\footnote{\url{https://github.com/InterDigitalInc/CompressAI/}} with the FFHQ training dataset.
Specifically, we adopt the $\lambda=\{1,1.5,2,2.5\}\times 10^{-4}$ to cover the similar bitrate range as ours.
In the evaluation of generative image compression methods, we perform a comparison with HiFiC~\cite{mentzer2020high}. 
%
%To achieve this, we retrain the PyTorch implemented model\footnote{\url{https://github.com/Justin-Tan/high-fidelity-generative-compression}} using the FFHQ training dataset, employing parameter values of $\lambda=\{192, 128,115,80\}$.
%
To achieve this, we retrain the PyTorch implemented model\footnote{\url{https://github.com/Justin-Tan/high-fidelity-generative-compression}} using the FFHQ training dataset, employing parameter values of $\lambda=\{1.92,1.28,1.15,0.8\}\times 10^{2}$.

\Paragraph{Evaluation Details.}
We perform facial landmark detection~\cite{wang2019adaptive} on the original CelebA-HQ dataset, the reconstructed images decoded by compared benchmarks, and the proposed Layer1.
Original images are used to detect ground-truth landmarks.
The landmark detection results are evaluated using the normalized mean error (NME)~\cite{wang2019adaptive} metric, which calculates the mean square error of the ground truth and the predicted landmark coordinates of each image as,
    \begin{equation}
        NME(p,\hat{p})=\frac{1}{N}\sum_{i=1}^{N}\frac{\|p_i-\hat{p}_i \|_2^2}{d},
    \end{equation}
where $p$ and $\hat{p}$ indicate the ground truth and the compared landmark coordinates of each image in the dataset, respectively.
$N$ denotes the number of landmarks of each image, $p_i$ and $\hat{p}_i$ is the $i-$th landmark coordinates in $p$ and $\hat{p}$.
$d$ is the normalized factor.
The lower value indicates that landmarks are detected more accurately.

Face parsing is based on the CelebAMask-HQ dataset~\cite{CelebAMask-HQ}, which has been manually annotated with 19 classes, including skin, nose, eyes, eyebrows, ears, mouth, lips, hair, and \etc.
%hat, eyeglasses, earring, necklace, neck, and clothing.
%
Thus, we apply the pre-trained face parsing model~\cite{CelebAMask-HQ} to the reconstructed dataset based on benchmarks and Layer1. 
Additionally, we report the prediction results over the original images for better comparisons.
We evaluate segmentation results based on frequency-weighted intersection over union (FwIoU)~\cite{garcia2017review}, which takes the IoU of each class and weights them according to their appearance frequency,
% \ie the area of overlap between the predicted segmentation and the ground truth divided by the area of their union, 
%
    \begin{equation}
        FwIoU=\frac{1}{\sum_{j=1}^{k}t_{j}}\sum_{j=1}^{k}t_{j}\frac{n_{jj}}{(n_{ij})+n_{ji}+n_{jj}}, i \neq j
    \end{equation}
where $n_{jj}$ is the total number of pixels that are both classified and labeled as class $j$ (true positives for class $j$), $n_{ij}$ is the number of pixels which are labeled as class $i$, but classified as class $j$ (false positives for class $j$), and $n_{ji}$ is the number of pixels which are labeled as class $j$, but classified as class $i$ (false negatives for class $j$), and $t_j$ is the total number of pixels labeled as class $j$.

%%%%%%%%%%%%%%%%%%%%%%%%%%

%%%%%%%%%%%%%%%%%%%%%%%%%%%%%
\begin{figure}[!t]
\centering
\includegraphics[width=0.8\linewidth]{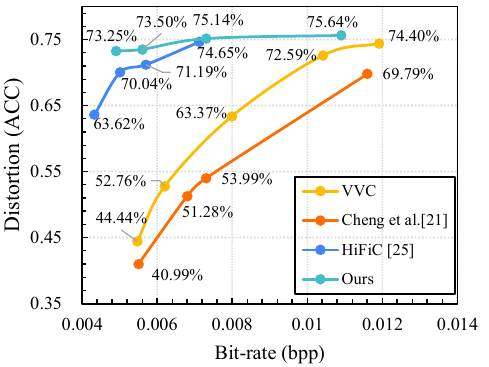}
\\
\caption{\textbf{The R-D performance of VVC, Cheng \etal~\cite{cheng2020image}, HiFiC~\cite{mentzer2020high}, and the middle layer on the CelebA-HQ dataset for facial identity recognition.}
Higher ACC values mean better recognition accuracy.
   %
%The proposed method can achieve much better accuracy, especially at ultra-low bit-rate. 
%
  }
    \label{fig:layer2-1}
\end{figure}
%%%%%%%%%%%%%%%%%%%%%%%%%%%%%%%%%%%%%%%%%%%%%%%%%%% 

\Paragraph{Evaluation Results Analysis.}
We present the R-D curve compared with VVC, Cheng \etal~\cite{cheng2020image} and HiFiC~\cite{mentzer2020high} at extremely low bitrate ($<0.01$ bpp) in \figref{RD-layer1}.
It can be clearly seen that our method achieves much higher accuracy at a similar bit-rate compared to VVC and  Cheng \etal~\cite{cheng2020image} in both face landmark detection and face parsing.
In particular, the NME and FwIoU of our method is $0.0286$ decrease, and $52.44\%$ increase than the VVC codec under bpp $\approx 0.0024$.
Compared to the original performance $89.31\%$, the proposed method achieves only $6\%-7\%$ accuracy loss under bpp $<0.006$, as shown in \figref{RD-layer1}(b).
\figref{layer1} presents a randomly selected example:
Under ultra-low bitrates, the VVC codec reconstructs images with blurry and blocking artifacts, and Cheng~\etal~\cite{cheng2020image} method produces over-smoothed decoded images, which results in inaccurate detection and parsing results.
In contrast, our method preserves the expression and shape of the original image, enabling commendable performance even without reconstructing detailed textures. 
While HiFiC~\cite{mentzer2020high} excels in generating abundant textures compared to other baselines, and demonstrates better texture consistency than the proposed basic layer, our approach outperforms by capturing essential semantic information necessary for landmark detection and parsing. As a result, our basic layer achieves superior performance, particularly under extremely low bitrates ($<0.003$ bpp).
Accordingly, the basic layer contains sufficient semantic information for landmark detection and parsing, whereas other codecs degrade such information under extremely low bitrates.

\subsection{Machine Vision: Compression Evaluation on the Middle Layer}
With transmitting an additional second layer of style vectors, the decoded image from the middle layer $\hat{\mathbf{x}}_2$ recovers more facial attributes textures.
In this section, we first perform the face identity recognition task.
Furthermore, we also evaluate on facial attribute prediction, which are not directly optimized in \secref{multi-task}.
This evidence indicates that our proposed method can be extended to other applications, in addition to the predefined vision task.

\Paragraph{Benchmarks.}
In this section, we also compare with the VVC, Cheng \etal \cite{cheng2020image}, and HiFiC~\cite{mentzer2020high}.
For VVC, we choose the QP from $\{55, 54, 52, 50, 49\}$ to cover the proposed method's bitrate range.
For Cheng \etal \cite{cheng2020image}, we retrain the model to cover the similar bitrate range as ours by using $\lambda=\{2,2.5,2.7,4.5\}\times 10^{-4}$.
In the case of HiFiC~\cite{mentzer2020high}, we employ $\lambda=\{4, 5.4, 6, 6.8\}\times 10$ to attain similar bitrates.

\Paragraph{Evaluation Details.}
For face identity recognition, we first process the CelebA-HQ dataset into $307$ identities according to the annotated labels.
Each identity has more than $15$ images and is split into training as well as testing sets according to the $8:2$ ratio, thus deriving $4263$ training and $1215$ testing images, respectively.
We use the original training set to fine-tune the pre-trained ResNet-18 model~\footnote{\UrlBigBreaks{https://github.com/ndb796/CelebA-HQ-Face-Identity-and-Attributes-Recognition-PyTorch}} and evaluate the testing accuracy (ACC) on the reconstructed images by benchmarks and the middle layer.
For facial attribute prediction, we adopt the pre-trained attribute classifier in \cite{lin2021anycost} to predict $40$ attributes on the CelebA-HQ dataset.
We compare prediction results on the original images and the compressed images to report the match rates (MR)~\cite{lin2021anycost}.

%%%%%%%%%%%%%%%%%%%%%%%%%%%%%%%%%%%%%%%%%%%%%%%%%%%%%%%
\begin{figure}[!t]
\centering
\includegraphics[width=1\linewidth]{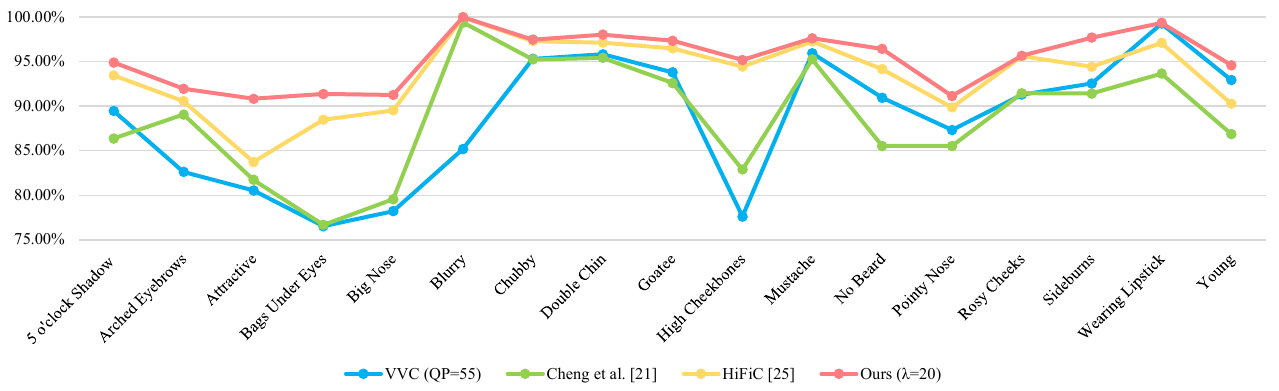}
\\
\caption{\textbf{Attribute matching rate comparisons with VVC, Cheng \etal~\cite{cheng2020image}, and HiFiC~\cite{mentzer2020high}.}
We show the advantage attribute MR of the proposed method against the benchmarks.
  }
    \label{fig:layer2-2}
\end{figure}
%%%%%%%%%%%%%%%%%%%%%%%%%%%%%%%%%%%%%%%%%%%%%%%%%%% 
%%%%%%%%%%%%%%%%%%%%%%%%%%%%%%%%%%%%%%%%%%%%%%%%%%%%%%%
\begin{figure}[!t]
\centering
\includegraphics[width=0.98\linewidth]{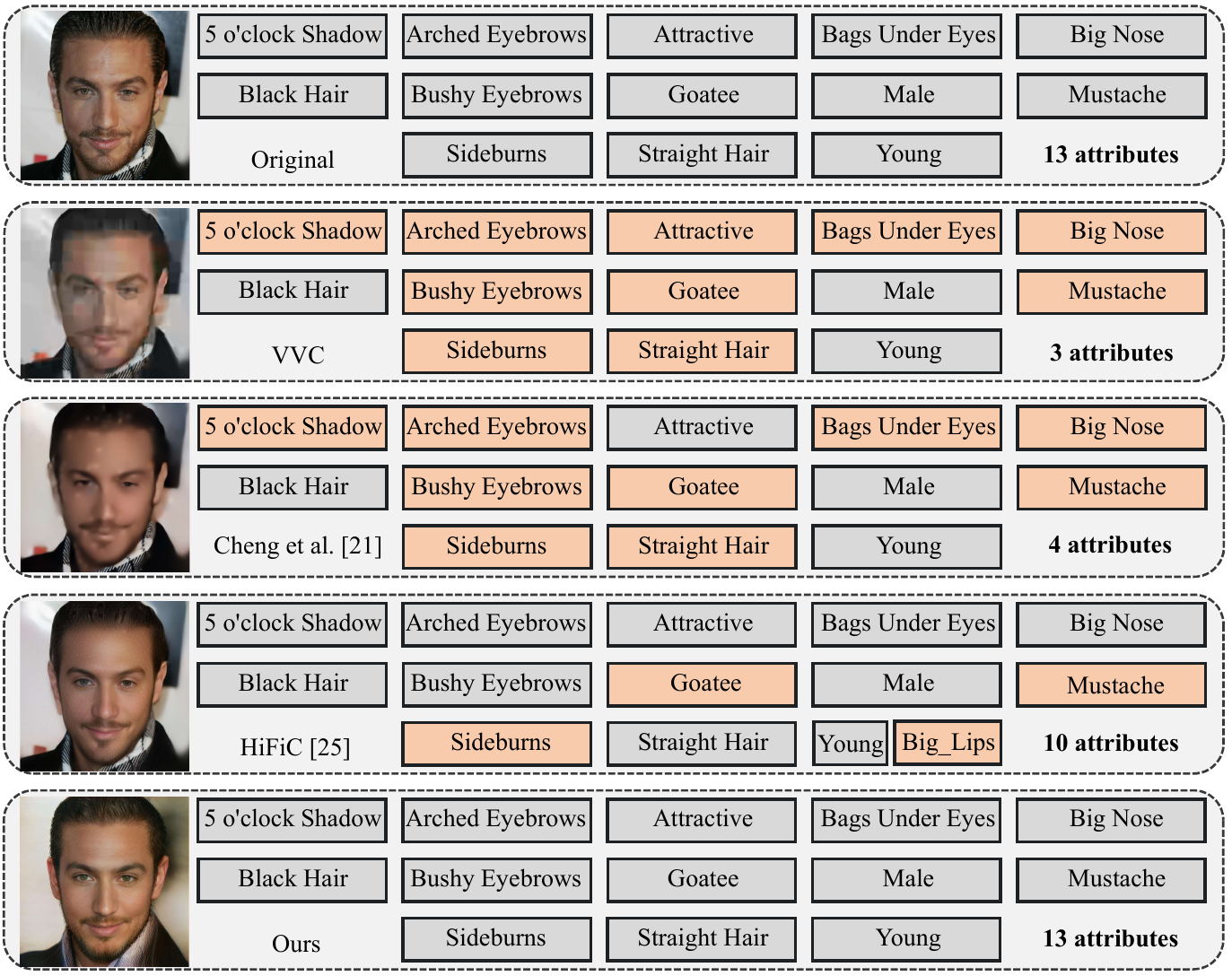}
\\
\caption{\textbf{Facial attribute prediction comparisons with VVC, Cheng \etal~\cite{cheng2020image}, and HiFiC~\cite{mentzer2020high}.}
Gray boxes indicate predicted attributes and orange boxes indicate missed attributes.
  }
    \label{fig:layer2-3}
\end{figure}
%%%%%%%%%%%%%%%%%%%%%%%%%%%%%%%%%%

\Paragraph{Evaluation Results Analysis.}
\figref{layer2-1} shows the R-D curve compared with VVC, Cheng \etal~\cite{cheng2020image} and HiFiC~\cite{mentzer2020high} in terms of face identity recognition accuracy \vs bitrates.
Owing to recovering most identity semantic attributes, our method achieves much higher recognition accuracy against VVC and Cheng \etal~\cite{cheng2020image}.
Specifically, the proposed middle layer achieves $28.81\%$ and $32.26\%$ gain against the VVC and Cheng \etal~\cite{cheng2020image} under bpp$<0.006$. 
Additionally, our method achieves better recognition accuracy than HiFiC~\cite{mentzer2020high} at bitrates below $0.008$ bpp.
For the facial attribute prediction task, we report MR curves in \figref{layer2-2} to better illustrate which attributes benefited from the middle layer.
It can be clearly observed that the proposed method can better preserve the attribute with much more texture details such as `Bags Under Eyes' and `High Cheekbones'.
The selected example in \figref{layer2-3} illustrates that the original image predicts $13$ facial attributes using the pre-trained classifier model.
However, the reconstructed image using VVC and Cheng \etal~\cite{cheng2020image} can only identify a small number of main attributes such as `Black Hair', `Young', and `Male'.
Despite the ability of HiFiC~\cite{mentzer2020high} to generate more texture than other baselines, it still suffers from degradation in attribute information, such as goatee, mustache, and sideburns, at extremely low bitrates.
In contrast, the proposed method accurately predicts all original attributes of the image.
As a result, the proposed method is able to reconstruct more semantic attribute textures, whereas other methods lose them during ultra-low bitrate compression.
Owing to its scalability, the proposed method allocates bit rates to identity and attribute texture information at the middle layer instead of other low-level texture details. 
This allocation strategy enhances the efficiency of our method.
%

%\vspace{-4 mm}

%%%%%%%%%%%%%%%%%%%%%%%%%%%%%%%%%%%%%%%%%
\begin{figure*}[!t]
\centering
\includegraphics[width=0.98\linewidth]{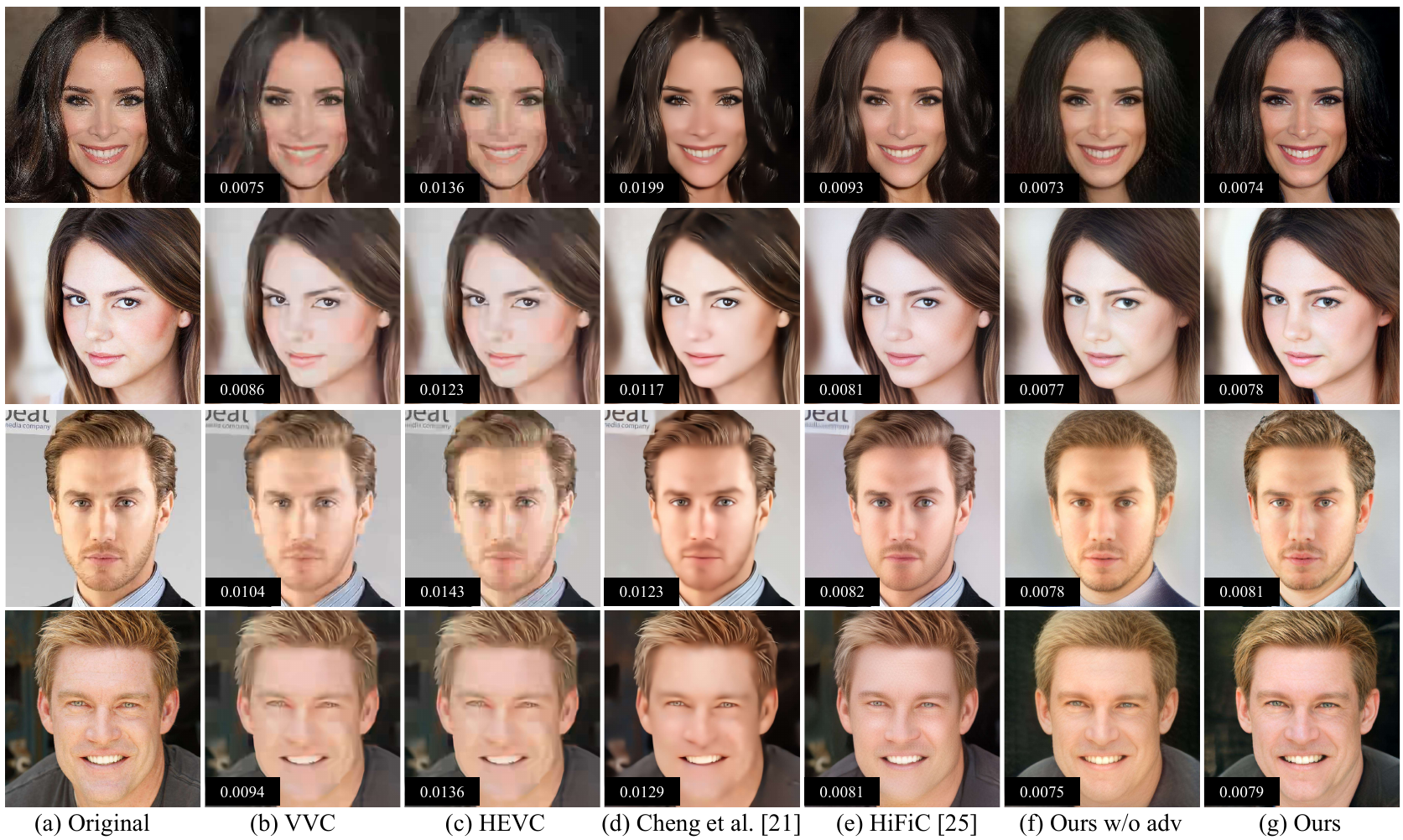}
\\
\caption{\textbf{Qualitative comparisons of VVC, HEVC, Cheng \etal~\cite{cheng2020image}, HiFiC~\cite{mentzer2020high}, and the enhanced layer on the CelebA-HQ dataset. }
(a) Input image. (b)-(c) Images compressed by VVC, HEVC using QP$=51$, and QP$=49$, respectively.
(d-e) Decoded images by Cheng \etal~\cite{cheng2020image} with $\lambda=5\times 10^{-4}$, and HiFiC~\cite{mentzer2020high} with $\lambda=37$.
(f)-(g) The proposed method w/o and w/ adversarial training at $\lambda=10$.
We present each decoded image's bit rate (bits per pixel/bpp) in the black boxes at the bottom.
  }
    \label{fig:layer3-3}
\end{figure*}
%%%%%%%%%%%%%%%%%%%%%%%%%%%%%%%%%%%%%%%%%
%%%%%%%%%%%%%%%%%%%%%%%%%%%%%%%%%%%%%%%%%%%%%%%%
\begin{figure*}[!t]
    \centering
    % \vspace{-3mm}
    \includegraphics[width=0.99\linewidth]{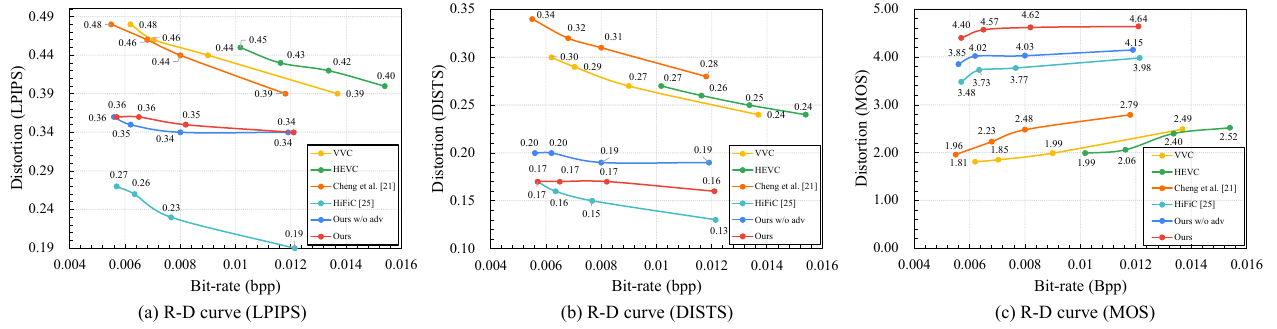}
   \caption{\textbf{The R-D performance of VVC, HEVC, Cheng \etal~\cite{cheng2020image}, HiFiC~\cite{mentzer2020high}, and the enhanced layer on the CelebA-HQ dataset.}
   (a)-(b) Lower LPIPS, DISTS values indicate better perceptual quality;
   (c) Higher MOS values indicate better human perceptual preference.}
    \vspace{-4mm}
    \label{fig:RD-layer3}
\end{figure*}

%%%%%%%%%%%%%%%%%%%%%%%%%%%%%%%%%%%%%%%%%%%%%%%%
\begin{figure}[!h]
    \centering
    % \vspace{-3mm}
    \includegraphics[width=0.99\linewidth]{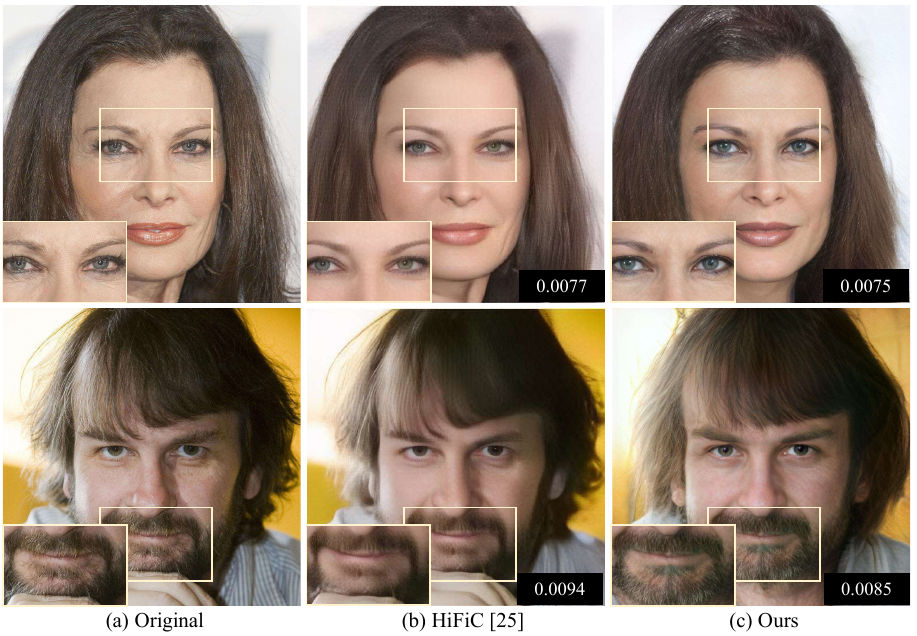}
   \caption{\textbf{Qualitative comparisons between reconstruction results of HiFiC~\cite{mentzer2020high} and the enhanced layer on the CelebA-HQ dataset.} 
   The bottom left corner exhibits an enlarged patch image.
Each decoded image's bit rate (bits per pixel/bpp) is shown in the black boxes at the bottom right.
}

    \label{fig:HiFiC-layer3}
\end{figure}
%%%%%%%%%%%%%%%%%%%%%%%%%%%%%%%%%%%%%%%%%%%%%%%%
%%%%%%%%%%%%%%%%%%%%%%%%%%%%%%%%%%%%
\subsection{Human Vision: Compression Evaluation on the Enhanced Layer}
%\vspace{-4 mm}
The third layer of style vectors captures more detailed low-level information, which makes the final decoded images from the enhanced layer reconstruct in a more promising quality for human vision.
We then evaluate the perceptual quality of the reconstructed image in this section.

\Paragraph{Benchmarks.}
In addition to VVC, we also compare with  HEVC~\cite{sullivan2012overview} by using all intra configuration of the reference software HM-16.25\footnote{\url{https://vcgit.hhi.fraunhofer.de/jvet/HM/-/tree/HM-16.25}}. 
To cover the bitrate range of proposed method, 
we choose the QP varying from $\{54, 53, 51, 48\}$ for VVC, and $\{51,50,49,48\}$ for HEVC, respectively.
For Cheng \etal~\cite{cheng2020image}, we use $\lambda=\{2,2.5,3,5\}\times 10^{-4}$ to adapt to the similar bitrate range as ours.
We employ $\lambda=\{2.4,3.7,4.8,5.4\} \times 10$ for HiFiC~\cite{mentzer2020high}.
Furthermore, we also develop an additional version by optimizing the proposed method without adversarial loss, denoted as ``Ours w/o adv'' for comparison.

\begin{table*}[t]
\centering
\caption{Quantitative evaluation of proposed framework scalability ($\lambda=15$).}
\label{tab:scalability}
\smallskip\noindent
\resizebox{0.7\linewidth}{!}{%
% \usepackage{booktabs}
% \usepackage{multirow}
\begin{tabular}{@{}cccccc@{}}
\toprule
\multicolumn{1}{c|}{\begin{tabular}[c]{@{}c@{}}Decoded\\ Layer\end{tabular}} & \multicolumn{1}{c|}{\begin{tabular}[c]{@{}c@{}}Bitrate\\ (BPP$\downarrow$)\end{tabular}} & \multicolumn{1}{c|}{\begin{tabular}[c]{@{}c@{}}Landmark Detection\\ (NME$\downarrow$)\end{tabular}}
&  \multicolumn{1}{c|}{\begin{tabular}[c]{@{}c@{}}Face Parsing\\ (FwIoU$\uparrow$)\end{tabular}} & 
\multicolumn{1}{c|}{\begin{tabular}[c]{@{}c@{}}Identity Recognition\\ (ACC$\uparrow$)\end{tabular}}&  
\multicolumn{1}{c}{\begin{tabular}[c]{@{}c@{}}Human Vision\\ (DISTS$\downarrow$)\end{tabular} }\\ \midrule
\multicolumn{1}{c|}{Basic Layer} & 
\multicolumn{1}{c|}{$0.0026$} & 
\multicolumn{1}{c|}{$0.0182$} & 
\multicolumn{1}{c|}{82.43\%} &                                      \multicolumn{1}{c|}{36.38\%} &   
\multicolumn{1}{c}{$0.25$}    
\\
 
 \multicolumn{1}{c|}{Middle Layer}& 
\multicolumn{1}{c|}{$0.0056$} & 
\multicolumn{1}{c|}{$0.0165$} & 
\multicolumn{1}{c|}{82.72\%} &                                      \multicolumn{1}{c|}{73.50\%} &   
\multicolumn{1}{c}{$0.20$} \\
 \multicolumn{1}{c|}{Enhanced Layer}& 
\multicolumn{1}{c|}{$0.0065$} & 
\multicolumn{1}{c|}{$0.0164$} & 
\multicolumn{1}{c|}{83.20\%} &                                      \multicolumn{1}{c|}{73.91\%} &   
\multicolumn{1}{c}{$0.17$} \\
\bottomrule
\vspace{-6mm}
\end{tabular}}
\end{table*}

\begin{table}[t]
\centering
\caption{Various Configurations for Ablation Study.}
\renewcommand\arraystretch{1.4}
\renewcommand\tabcolsep{1.6 pt}
\begin{tabular}{lccccc} 
\toprule
Method & A & B & C & D & E \\
\hline
Conv Hyperprior& \cmark & $\times$ & $\times$ & $\times$ & $\times$  \\
Layer-Wise Hyper-Transformer & $\times$  & \cmark & \cmark & \cmark& \cmark \\
%Hyper-Transformer& \\
Cross-Layer Entropy Transformer& $\times$  & $\times$ &\cmark & \cmark & \cmark\\
Multi-Task Scalable R-D Optimization &$\times$&$\times$ &$\times$  & \cmark & \cmark\\
Adversarial Training & $\times$  & $\times$ &$\times$  & $\times$  & \cmark\\
\bottomrule 
\vspace{-6mm}
\label{tab:configs}
\end{tabular}
\end{table}

\begin{table}[!t]
\caption{ Quantitative results on configurations of \tabref{configs} ($\lambda=10$).}
\renewcommand\arraystretch{1.3}
\renewcommand\tabcolsep{1.4 pt}
 \label{tab:ablation-studies}
 \centering
 \begin{tabular}{l cccccc}
 \toprule
 Layer & Metrics & A & B & C & D & E \\
 \midrule
 \multirow{3}{*}{Basic Layer}&  BPP$\downarrow$ & $0.0032$ & $0.0030$ & $0.0027$ & $0.0036$ & $0.0036$ \\
 & NME$\downarrow$& $0.0229$ & $0.0223$ & $0.0229$ & $0.0162$ & $0.0175$\\
 & FwIoU$\uparrow$ & $78.32\%$ & $78.78\%$ & $79.10\%$ & $81.89\%$ & $82.63\%$\\
 \cline{1-7}
 \multirow{2}{*}{Middle Layer}&  BPP$\downarrow$ & $0.0084$ & $0.0081$ & $0.0075$ & $0.0073$ &$0.0073$ \\
 & ACC $\uparrow$ & $61.32\%$ & $62.30\%$ & $60.99\%$ & $71.19\%$ & $75.14\%$ \\
\cline{1-7}
 \multirow{3}{*}{Enhanced Layer}&  BPP$\downarrow$ & $0.0094$ & $0.0092$ & $0.0085$ & $0.0080$ & $0.0082$ \\
  & LPIPS$\downarrow$ &$0.34$ & $0.34$ & $0.34$ & $0.34$ & $0.35$ \\
 & DISTS$\downarrow$ &$0.20$ & $0.19$ & $0.20$ & $0.19$ & $0.17$  \\
 \bottomrule 	
 \end{tabular}
 \vspace{-0.1cm}
\end{table}

\Paragraph{Evaluation Details.}
Firstly, we adopt recent attempts towards perceptual quality assessment, which are better aligned with human perception.
In particular, the learned perceptual image patch similarity (LPIPS)~\cite{zhang2018unreasonable} and the Ddeep image structure and texture similarity (DISTS)~\cite{ding2020image} metrics are adopted.
While the former calculates distance within AlexNet embedding feature space, the latter uses deep features to unify structure and texture similarity for better-capturing texture perception.
Furthermore, to better understand the human vision perception, we conduct a user study and apply double stimulus image quality assessment to collect the subjective image quality scores varying from $\{1,2,3,4,5\}$ which corresponds to the subjective quality level of $\{$``Bad'',``Poor'',``Fair'',``Good'', ``Excellent''$\}$~\cite{bt2002methodology}.
This study evaluates $24$ images randomly selected from the CelebA-HQ dataset on four compression levels of each method. %
We present the original and reconstructed images decoded using compared methods for each test.
Participants are required to view the screen at a resolution of 1080P or higher on their viewing device.
We collect the answers from $25$ non-expert participants. 
Then the Mean Opinion Score (MOS)
is used to measure the human perceptual preference as,
    \begin{equation}
        \mathrm{MOS}_{i}=\frac{1}{MN}\sum_{m=1}^{M}\sum_{n=1}^{N},
    \end{equation}
where for each compressed level $i$, the $N$ and $M$ is the total number of participants and testing images.

\Paragraph{Evaluation Results Analysis.}
For a better understanding of how the compression performs towards human perception, we plot the R-D curve at extremely low bitrate using LPIPS, DISTS, and MOS as distortion metrics in \figref{RD-layer3}.
\figref{layer3-3} shows that the decoded images from the enhanced layer generate more perceptual pleasing facial semantics under similar bitrates, thereby realizing much higher LPIPS and DISTS values than the traditional codecs as well as Cheng \etal~\cite{cheng2020image} in \figref{RD-layer3}(a)-(b).
\figref{RD-layer3}(c) also illustrates that the blurriness and blocking artifacts have negative effects towards human perceptual preference.
The DISTS measurement favors abundant texture, so we achieve higher DISTS values with adversarial loss than our version without adversarial loss.
Moreover, our proposed method, which emphasizes facial texture over background detail preservation, yields less favorable results in terms of LPIPS and DISTS metrics when compared to HiFiC~\cite{mentzer2020high}.
Nevertheless, HiFiC~\cite{mentzer2020high} exhibits severe checkerboard artifacts on facial regions at extremely low bitrates, whereas our approach presents a more natural and enriched facial texture, as illustrated in \figref{HiFiC-layer3}.
Consequently, our proposed method outperforms HiFiC~\cite{mentzer2020high} in subjective user studies, achieving a higher MOS score in \figref{layer3-3}(c).

\subsection{Scalability Analysis.}
In this section, we demonstrate the scalable coding capabilities of the proposed framework.
\tabref{scalability} presents the quantitative evaluation of three-layered decoding results using the model with $\lambda=15$.
With more bits decoded, the accuracy rate of landmark detection, segmentation, identity recognition, and human perceptual quality all increase, thus verifying the scalability of the proposed framework.
%need refine
Furthermore, it can be clearly observed that each layer has achieved promising performance on its corresponding vision tasks.
The basic layer, for instance, achieves $82.43\%$ FwIoU score for face parsing, while the middle layer gets an ACC metric of $73.50\%$ for identity recognition, with only a drop in performance of $0.77\%$ and $0.41\%$ against the enhanced layer respectively, which indicates that the proposed scheme contains enough semantic information at each layer to perform specific vision tasks.

%%%%%%%%%%%%%%%%%%%%%%%%%%%%%%%%%%%%%%%%%%%%%%%%
\begin{figure}[!t]
    \centering
    % \vspace{-3mm}
    \includegraphics[width=0.99\linewidth]{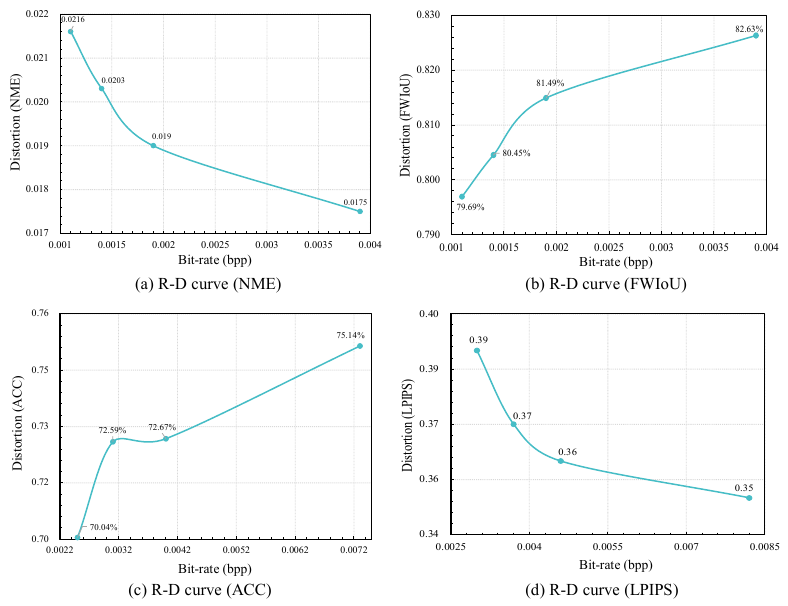}
   \caption{\textbf{The R-D performance on varying tasks of different layers.}
      (a-b) The basic layer: lower NME values indicate better landmark detection accuracy, and higher FWIoU values mean better parsing accuracy;
      (c) The middle layer: higher ACC values mean better recognition accuracy;
  (d) The enhanced layer: lower LPIPS values indicate better perceptual quality.}
    %\vspace{-4mm}
    \label{fig:RD-lower}
\end{figure}
%%%%%%%%%%%%%%%%%%%%%%%%%%%%%%%%%%%%%%%%%%%%%%%

\subsection{Ablation Study.}
To gain a better understanding of each module and training strategies of the proposed method, we conduct a comprehensive ablation study by \emph{sequentially incorporating our proposed methodologies}, thereby systematically assessing their individual impacts and contributions, as illustrated in \tabref{configs}.
%
%\Paragraph{Proposed network modules.}
\secref{entropy-model} introduces two transformer entropy modules to exploit the correlations of three layered style vectors.
By leveraging the multi-head self-attention module, the layer-wise hyper-transformer can exploit both cross-layered style vectors and cross-channel redundancies.
The cross-layer entropy transformer models the condition relation between the decoded and the current layers.
In \secref{multi-task}, we propose a multi-task scalable R-D optimization method by simultaneously optimizing three-layered decoded images under a controlled bitrate constraint.
We first analyze the efficiency of two proposed network modules.
Then, we discuss how the proposed training strategies influence the final results.

\Paragraph{Proposed network modules.}
We initially introduce a variant, namely method A, which employs 1$\times$1 convolutions similar to the approach in~\cite{chang2021thousand} to leverage cross-channel redundancy. 
Subsequently, we utilize the layer-wise hyper-transformer to develop method B. 
To confirm the positive impact of the multi-head self-attention module, we compare method B with method A, as presented in \tabref{ablation-studies}. 
The utilization of multi-head attention effectively exploits cross-layered style vector redundancies, leading to greater bitrate savings and improved performance for all three-layered decoded images.
Next, we incorporate a cross-layer entropy transformer in method C and compare it with method B. By considering information from previously decoded layers, method C achieves more accurate estimations, resulting in a more substantial reduction in bitrate compared to method B.

\Paragraph{Proposed training strategies.}
To explore the impact of multi-task scalable R-D optimization, we compare method C with method D. 
While method D's first layer of style vectors consumes more bitrates compared to method C, it achieves an overall reduction in bitrate. 
Moreover, machine vision tasks performed on images from the basic and middle layers exhibit notable improvements. 
Additionally, the adversarial training of method E enhances textures in enhanced layer images, resulting in a better DISTS score when compared to method D.
%

%
%%%%%%%%%%%%%%%%%%%%%%%%%%%%%%%%%%%%%%%%%%%%%%%%
\begin{figure}[!t]
    \centering
    % \vspace{-3mm}
    \includegraphics[width=0.99\linewidth]{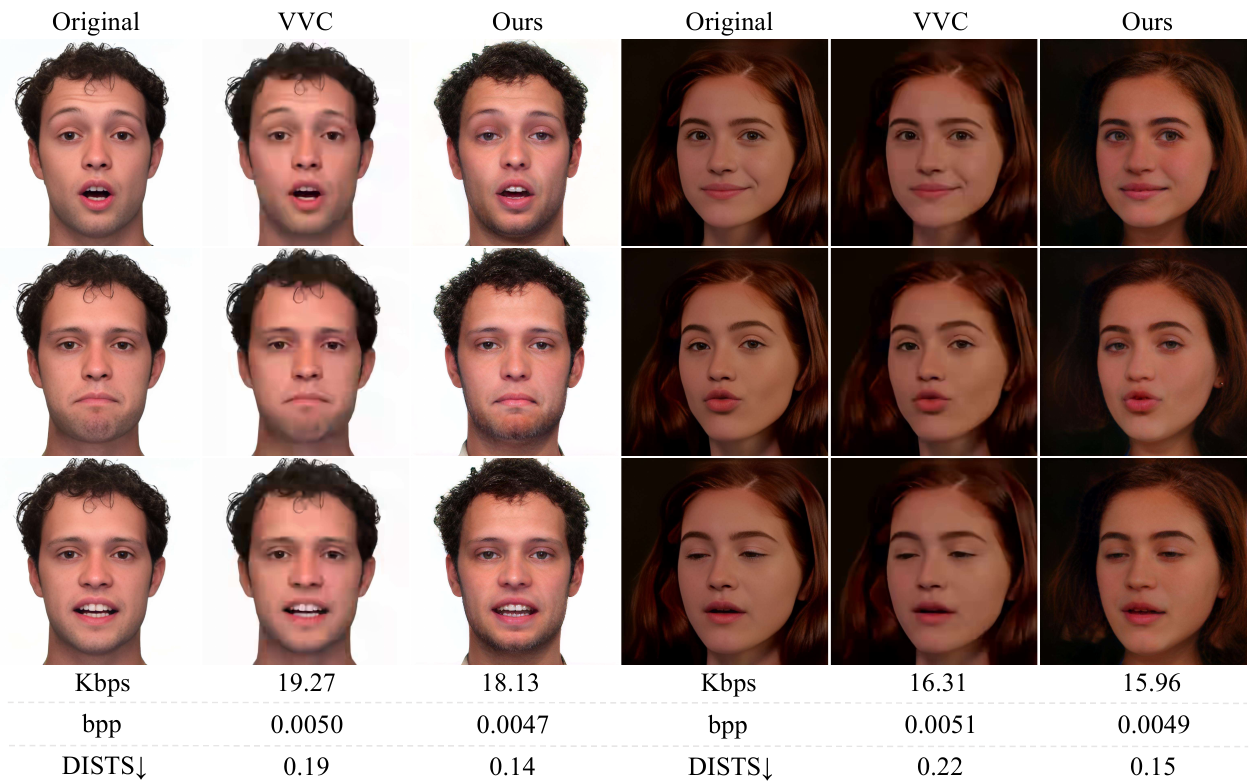}
   \caption{\textbf{The qualitative reconstruction results of the proposed method on facial video of RAVDESS~\cite{livingstone2018ryerson} and CelebV-HQ~\cite{zhu2022celebv} dataset.}
Lower DISTS values indicate the presence of abundant textures. 
In comparison to VVC, the proposed method achieves improved visual quality while reducing bitrates.
}
    \label{fig:video}
\end{figure}
%%%%%%%%%%%%%%%%%%%%%%%%%%%%%%%%%%%%%%%%%%%%%%%%
%%%%%%%%%%%%%%%%%%%%%%%%%%%%%%%%%%%%%%%%%%%%%%%%%%%%%%%
\begin{figure}[!t]
\centering
\includegraphics[width=1.0\linewidth]{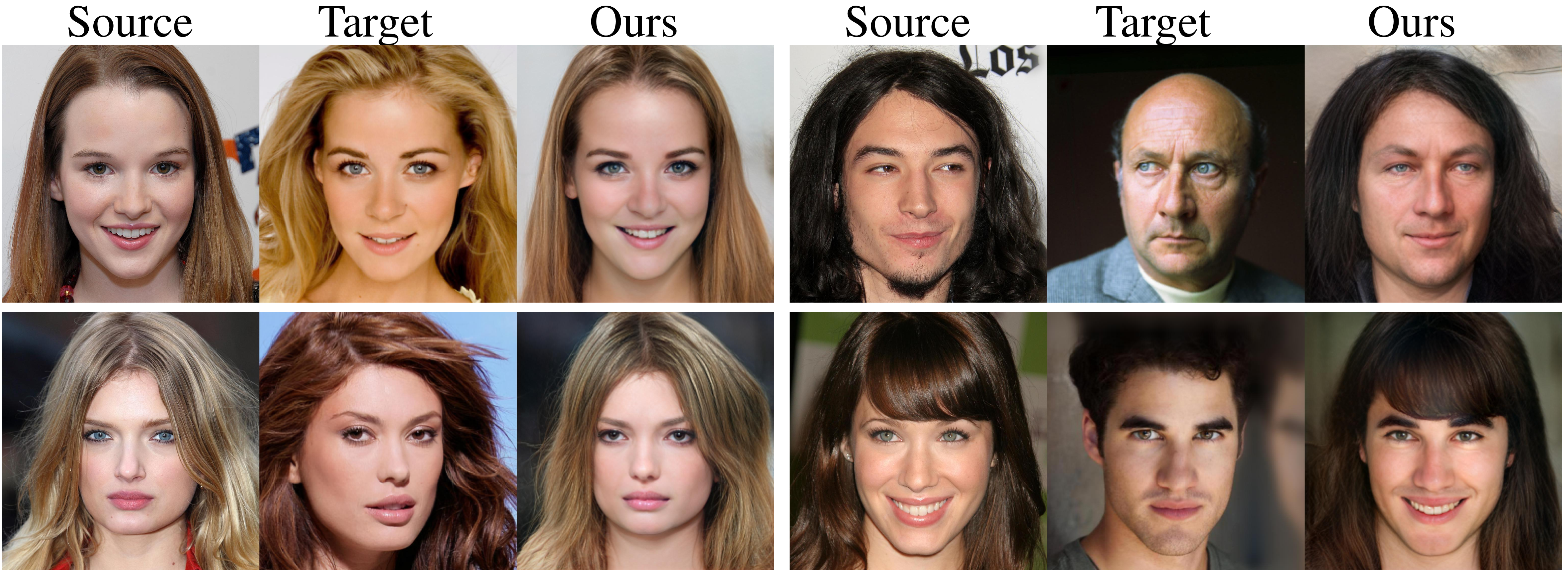}
\\
\caption{\textbf{Illustrations of face swap.}
We reserve the first and third layer's style vectors of the source image and swap the second layer's style vectors of the target image to produce the face swap effects.
  }
    \label{fig:face-swap}
\end{figure}
%%%%%%%%%%%%%%%%%%%%%%%%%%%%%%%%%%%%%%%%%%%%%%%%%%% 
%\vspace{-2mm}
\subsection{Discussions}
\label{sec:applications}
This section begins by discussing the considerably lower bitrates performance of the proposed compression framework. Next, it introduces two applications: facial video compression and face swapping. Finally, the limitations are summarized.

\noindent
\Paragraph{Performance on significantly lower bitrates.} 
We perform additional experiments at substantially lower bitrates by adjusting the parameter $\lambda$.
The resulting R-D curve, covering a range of $\lambda=\{10, 30, 50, 70\}$ for various tasks on each layer, is depicted in \figref{RD-lower}.
These comprehensive experiments clearly reveal that task performance exhibits significant variations at much lower bitrates, providing strong evidence that $\lambda$ effectively balances the rate-performance trade-off across layers.

\noindent
\Paragraph{Applications on facial video compression.} 
By disentangling the three-layered representations, our proposed method readily extends its application to face video compression.
Specifically, we capture and represent motion information using the basic layer representation, which preserves landmark and pose information.
Consequently, we strategically employ the middle and enhanced layer representations of key frames while transmitting only the basic layer for non-key frames, efficiently reducing temporal redundancy.
The proposed method is evaluated on  the RAVDESS~\cite{livingstone2018ryerson} and CelebV-HQ~\cite{zhu2022celebv} datasets without additional training on any video dataset.
We adopt a frame structure consisting of three frames, with one frame designated as a keyframe and the remaining frames as non-keyframes. 
For VVC, we employ the low-delay mode of VTM-11.0.
As demonstrated in \figref{video}, our method demonstrates superior visual textures compared to VVC, which exhibits noticeable blurring artifacts at similar bitrates. This provides compelling evidence of the effectiveness of our approach in facial video coding.
%}

\noindent
\Paragraph{Applications on Face Swapping.}
Attributed to the disentanglement of three-layered representations, the proposed coding framework can be applied to facial swapping without additional training.
As shown in \figref{face-swap}, it can easily swap face images by replacing the second layer of style vectors for the target identity while preserving the first and third layers of the source image.
By swapping out the selected face images' identities from the database, we can render any user's preferred face images without decoding.
%

% %%%%%%%%%%%%%%%%%%%%%%%%%%%%%%%%%%%%%%%%%%%%%%%%%%%%%%%
% \begin{figure}[!t]
% \centering
% \includegraphics[width=1.0\linewidth]{Image/Image-Manipulation/video.pdf}
% \\
% \caption{\textbf{Video compression.}
% %
%   }
%     \label{fig:video}
% \end{figure}
% %%%%%%%%%%%%%%%%%%%%%%%%%%%%%%%%%%%%%%%%%%%%%%%%%%% 

% \Paragraph{Video Compression.}
% The proposed method can be easily extended into the video compression, as demonstrated in \figref{video}. 
% %
% Note that we do not train on any video.

\Paragraph{Limitations.}
There are several limitations to the proposed codec framework.
Due to the characteristics of StyleGAN prior, the proposed framework is designed primarily for specific-domain compression, such as facial images, and cannot be applied directly to other image domains without additional training.
Furthermore, although our framework yields promising results with ultra-low bitrate compression, the styleGAN's upper-performance limit prevents it from achieving better results with a much higher bitrate.

\section{Conclusions}
\label{sec:conclusions}
In this work, we present one of the first efforts to leverage layer-wise style vectors of StyleGAN prior as compact visual data representations, assigning their hierarchical semantic information into the basic, middle, and enhanced layers to progressively support machine analysis and human perception.
The novel properties of the proposed scheme lie in the hierarchical semantic information assignments of three-layered representations, cross-layer correlation reduction by the layer-wise scalable entropy transformer, and the elaborately designed multi-task scalable optimization strategies, enabling efficient human-machine collaborative compression.
We demonstrate the superior performance of the proposed scheme on face image compression: corresponding vision tasks on each scalable layer can be performed more effectively against the latest traditional and learning-based compression paradigms.
Overall, the proposed scheme advances the field of image/video coding research by showing how semantic representations derived from generative prior could offer new insights to develop efficient scalable coding schemes for human-machine collaborative visions.

\bibliographystyle{IEEEbib}
\bibliography{refs}

%%%%%%%%%%%%%%%%%%%%portrait_photos

\begin{IEEEbiography}  
[{\includegraphics [width=1in,height=1.25in,clip,keepaspectratio] {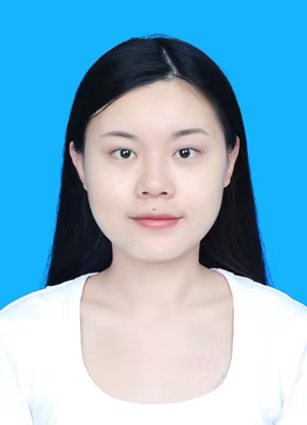}}]
{Qi Mao}
(Member, IEEE) received the B.S. degree in
digital media technology from the Communication
University of China in 2016 and the Ph.D. degree
in EECS from the Institute of Digital Media, Peking
University, in 2021. She was a visiting student at the
Vision and Learning Lab, University of California,
Merced, CA, USA, in 2019. 
Currently, She is an
Associate Professor with the School of Information
and Communication Engineering and the State Key
Laboratory of Media Convergence and Communication, Communication University of China. Her
research interests include intelligent
image/video compression and AIGC.
\end{IEEEbiography}

\begin{IEEEbiography}  
[{\includegraphics [width=1in,height=1.25in,clip,keepaspectratio] {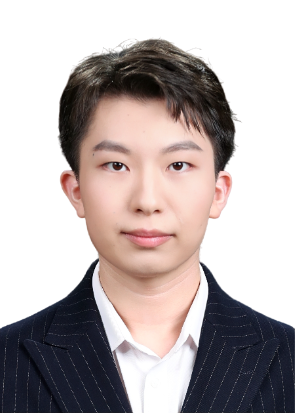}}]
{Chongyu Wang}
received the B.Eng. degree in Information Engineering from the Beijing Technology and Business University in 2017 and the M.Eng. degree in Electronic Information from the Communication University of China in 2023, His research interests include image/video compression and deep generative models.
\end{IEEEbiography}

\begin{IEEEbiography}  
[{\includegraphics [width=1in,height=1.25in,clip,keepaspectratio] {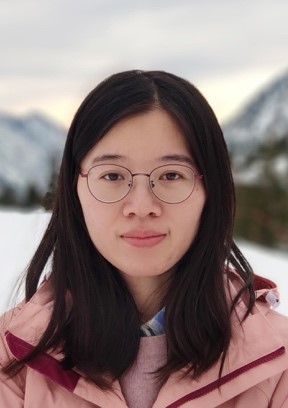}}]
{Meng Wang}
(Member, IEEE) received the B.S. degree in electronic information engineering of Honors Program from China Agricultural University, Beijing, China, in 2015, the M.S. degree in computer application technology from Peking University, Beijing, China, in 2018, and the Ph.D. degree in computer science from the City University of Hong Kong, Hong Kong, China, in 2021. She is currently a postdoc with the Department of Computer Science, City University of Hong Kong. She was a research intern in Bytedance Inc., since 2017. Her research interests include data compression and image/video coding.
\end{IEEEbiography}

\begin{IEEEbiography}  
[{\includegraphics [width=1in,height=1.25in,clip,keepaspectratio] {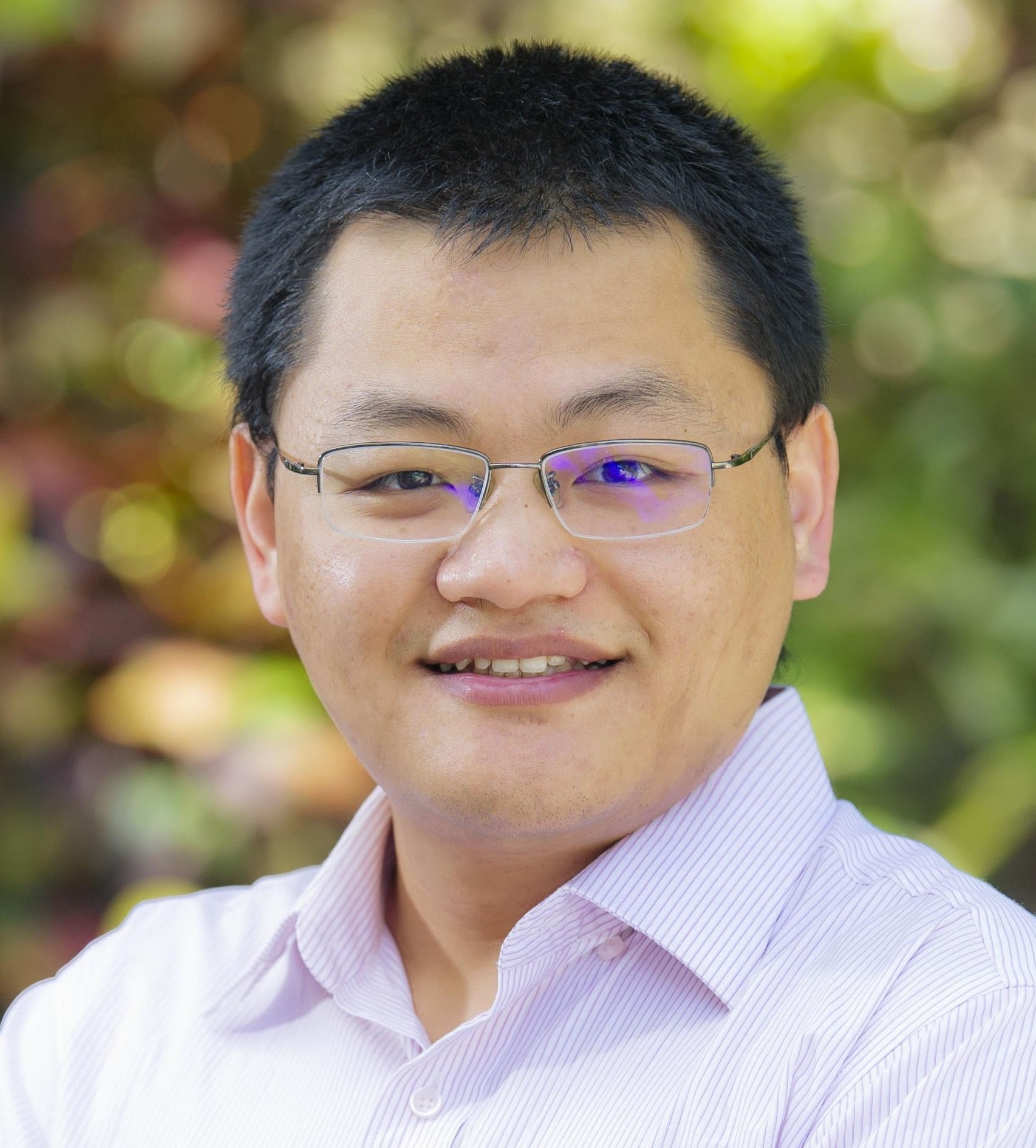}}]
{Shiqi Wang}
(Senior Member, IEEE) received the Ph.D. degree in computer application technology from Peking University in 2014. He is currently an Associate Professor with the Department of Computer Science, City University of Hong Kong. He has proposed more than 70 technical proposals to ISO/MPEG, ITU-T, and AVS standards. He authored or coauthored more than 300 refereed journal articles/conference papers, including more than 100 IEEE Transactions. His research interests include video compression, image/video quality assessment, video coding for machine, and semantic communication. He received the Best Paper Award from IEEE VCIP 2019, ICME 2019, IEEE Multimedia 2018, and PCM 2017. His coauthored article received the Best Student Paper Award in the IEEE ICIP 2018. He served or serves as an Associate Editor for IEEE TIP, TCSVT, TMM, TCyber, Access, and APSIPA Transactions on Signal and Information Processing.
\end{IEEEbiography}

\begin{IEEEbiography}  
[{\includegraphics [width=1in,height=1.25in,clip,keepaspectratio] {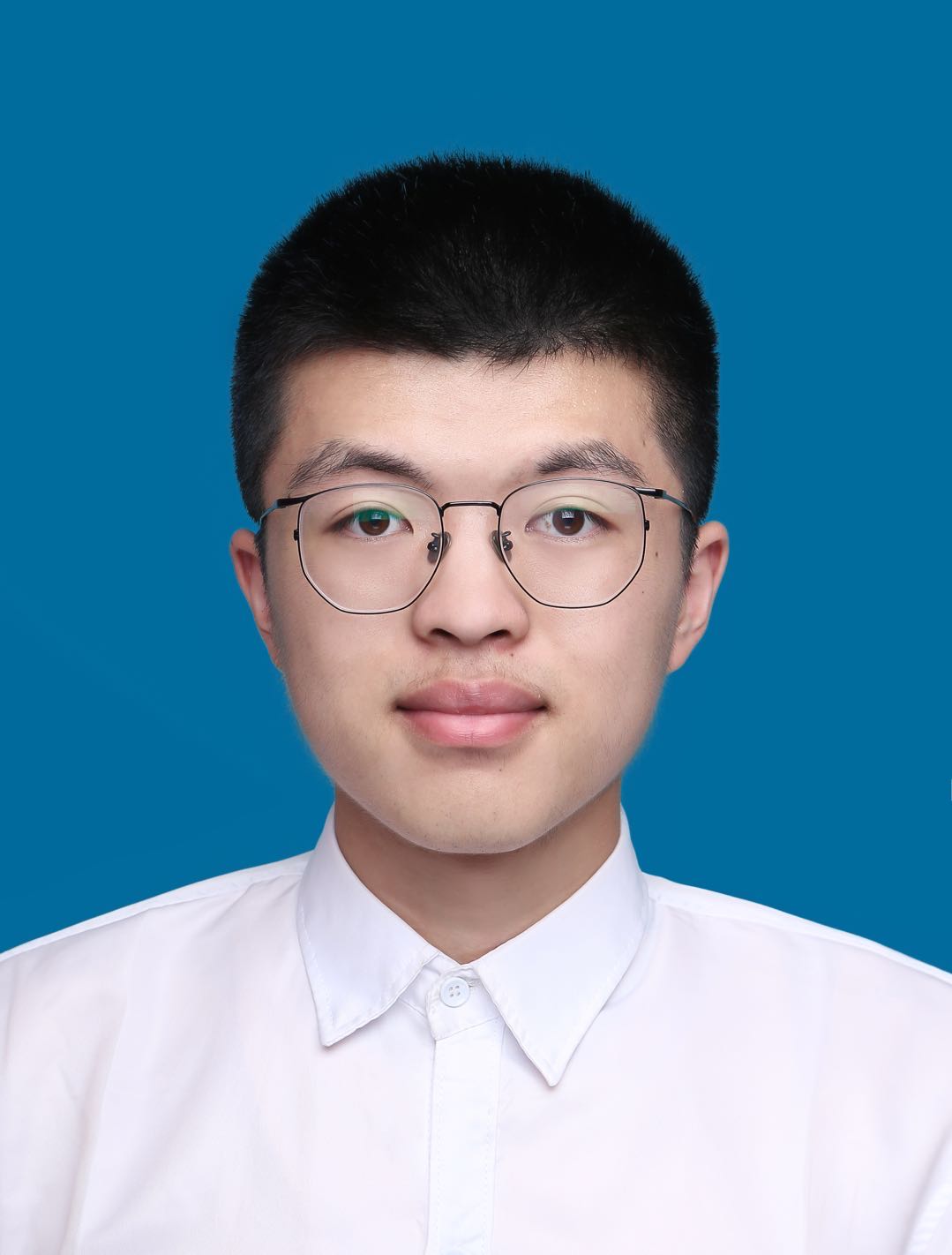}}]
{Ruijie Chen}
is studying for the B.E. degree in digital media technology at Communication University of China, Beijing, China. His research interests include image synthesis and image/video compression.
\end{IEEEbiography}

\vspace{-7cm}

\begin{IEEEbiography}  
[{\includegraphics [width=1in,height=1.25in,clip,keepaspectratio] {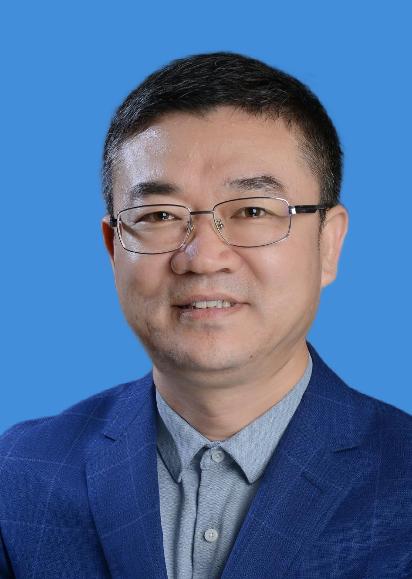}}]
{Libiao Jin}
(Member, IEEE) received the B.S. degree in electronic information engineering from Minzu University of China in 2000, and the M.S. and Ph.D. degrees from Communication University of China in 2003 and 2008. He is currently a Professor with the School of Information and Communication Engineering, Communication University of China, Beijing, China. He is the author of three books, more than 100 articles. His research interests mainly include artificial intelligence and multimedia communication.
\end{IEEEbiography}

\vspace{-7cm}

\begin{IEEEbiography}  
[{\includegraphics [width=1in,height=1.25in,clip,keepaspectratio] {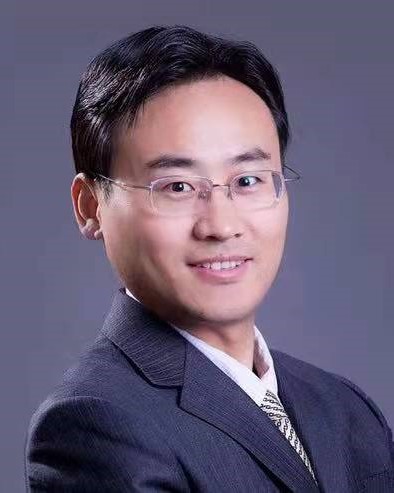}}]
{Siwei Ma}
(Fellow, IEEE) received the B.S. degree from Shandong Normal University, Jinan, China, in 1999, and the Ph.D. degree in computer science from the Institute of Computing Technology, Chinese Academy of Sciences, Beijing, China, in 2005. He held a postdoctoral position with the University of Southern California, Los Angeles, CA, USA, from 2005 to 2007. He joined the School of Electronics Engineering and Computer Science, Institute of Digital Media, Peking University, Beijing, where he is currently a Professor. He has authored over 300 technical articles in refereed journals and proceedings in image and video coding, video processing, video streaming, and transmission. He served/serves as an Associate Editor for the IEEE Transactions on Circuits and Systems for Video Technology and the Journal of Visual Communication and Image Representation.
\end{IEEEbiography}

\end{document}